\documentclass[10pt,journal,compsoc]{IEEEtran}
\usepackage{xr}
\usepackage[lined,ruled,commentsnumbered]{algorithm2e}
\usepackage{graphicx,url,subfigure,multirow,booktabs,amsmath,amssymb,threeparttable,bm,color}
\usepackage{makecell}
\ifCLASSOPTIONcompsoc
  \usepackage[nocompress]{cite}
\else
  \usepackage{cite}
\fi

\allowdisplaybreaks

\begin{document}

\title{BoostTree and BoostForest\\ for Ensemble Learning}

\author{Changming~Zhao, Dongrui~Wu,~\IEEEmembership{Fellow,~IEEE,}
        Jian~Huang,~\IEEEmembership{Senior~Member,~IEEE,}
        Ye~Yuan,~\IEEEmembership{Senior~Member,~IEEE,}
        Hai-Tao~Zhang,~\IEEEmembership{Senior~Member,~IEEE,}\\
        Ruimin~Peng and  Zhenhua~Shi
\IEEEcompsocitemizethanks{\IEEEcompsocthanksitem Key Laboratory of the Ministry of Education for Image Processing and Intelligent Control, School of Artificial Intelligence and Automation, Huazhong University of Science and Technology, Wuhan, China.
	C. Zhao, D. Wu and R. Peng are also with Shenzhen Huazhong University of Science and Technology Research Institute, Shenzhen, China.
H-T Zhang is also with the Autonomous Intelligence Unmanned Systems Engineering Research Center of Ministry of Education of China,
and the State Key Lab of Digital Manufacturing Equipment and Technology, Wuhan, China.
\IEEEcompsocthanksitem Dongrui Wu is the corresponding author. E-mail: drwu@hust.edu.cn.}}

\markboth{IEEE Transactions on Pattern Analysis and Machine Intelligence}%
{Zhao \MakeLowercase{\textit{et al.}}: BoostTree and BoostForest for Ensemble Learning}

\IEEEtitleabstractindextext{%
\begin{abstract}
Bootstrap aggregating (Bagging) and boosting are two popular ensemble learning approaches, which combine multiple base learners to generate a composite model for more accurate and more reliable performance. They have been widely used in biology, engineering, healthcare, etc. This paper proposes BoostForest, which is an ensemble learning approach using BoostTree as base learners and can be used for both classification and regression. BoostTree constructs a tree model by gradient boosting. It increases the randomness (diversity) by drawing the cut-points randomly at node splitting. BoostForest further increases the randomness by bootstrapping the training data in constructing different BoostTrees. BoostForest generally outperformed four classical ensemble learning approaches (Random Forest, Extra-Trees, XGBoost and LightGBM) on 35 classification and regression datasets. Remarkably, BoostForest tunes its parameters by simply sampling them randomly from a parameter pool, which can be easily specified, and its ensemble learning framework can also be used to combine many other base learners.
\end{abstract}

}

\maketitle

\IEEEdisplaynontitleabstractindextext
\IEEEpeerreviewmaketitle

\IEEEraisesectionheading{\section{Introduction}\label{sec:introduction}}

\IEEEPARstart{E}{nsemble} learning \cite{zhou2012ensemble,hastie2009elements} trains multiple base learners to exploit the relationship between a set of covariates (features) and a response (label), and then combines them to produce a strong composite learner with better generalization performance. It has been successfully used in biology \cite{Wang2019,Sun2020,Cao2020b}, engineering \cite{Paisitkriangkrai2016,Yang2020,Li2021}, healthcare \cite{Kim2020,Agius2020}, etc.

For example, in biology, Wang et al. \cite{Wang2019} used an ensemble of neural networks to emulate mechanism-based biological models. They found that the ensemble is more accurate than an individual neural network, and the consistency among the individual models can indicate the error in prediction. In Moon exploration, Yang et al. \cite{Yang2020} used ensemble transfer learning and Chang'E data for automatic Moon surface impact crater detection and age estimation. They successfully identified 109,956 new craters from 7,895 training samples. In healthcare, Agius et al. \cite{Agius2020} developed a Chronic Lymphocytic Leukemia (CLL) treatment-infection model, an ensemble of 28 machine learning algorithms, to identify patients at risk of infection or CLL treatment within two years of diagnosis.

One of the most popular algorithms for constructing the base learners is the decision tree \cite{friedman2001greedy,chen2016xgboost,breiman1984classification}. Two common approaches for constructing the composite learner are bootstrap aggregating (Bagging) and boosting.

Bagging \cite{breiman1996bagging} connects multiple base learners in parallel to reduce the variance of the ensemble. Each base learner is trained using the same learning algorithm on a bootstrap replica, which draws $N$ (the size of the original training set) samples with replacement from the original training set. The outputs of these base learners are then aggregated by majority voting (for classification) or averaging (for regression) to obtain the final output. To achieve robust performance, the base learners in an ensemble should be both accurate and diverse \cite{kuncheva2003measures,tang2006analysis,brown2005diversity}.

There are many approaches to increase the accuracy of base learners in an ensemble. Combining the advantages of tree models and linear models can greatly improve the model's learning ability, which is the main idea of model trees. M5 \cite{quinlan1992learning} constructs a linear regression function at each leaf to approximate the target function for high fitting ability. When a new sample comes in, it is first sorted down to a leaf, then the linear model at that leaf is used to predict its output. M5P (aka M5$'$) \cite{wang1997induction} trains a linear model at each node of a pruned tree to reduce the risk of over-fitting. More sophisticated regression algorithms, e.g., Ridge Regression (RR) \cite{hastie2009elements}, Extreme Learning Machine (ELM) \cite{huang2006extreme}, Support Vector Regression (SVR) \cite{drucker1997support}, and Neural Network \cite{Jain1996}, have rarely been used in model trees. A possible reason is that they have some hyper-parameters to tune, e.g., the regularization coefficient and the number of nodes in the hidden layers. It is an NP-complete problem to simultaneously determine the structure of the model tree and the parameters of each node model \cite{HYAFIL197615}. A common approach is cross-validation, but it is very time-consuming. Thus, it is desirable to develop a strategy that can make the model tree more compatible with these more sophisticated (and hence potentially better performing) regression models.

There are also many approaches to increase the diversity of base learners in an ensemble, which can be divided into three categories \cite{Zhou2018}: 1) Sample-based strategies, which train each base learner on a different subset of samples, and thus are scalable to big data. For example, Bagging uses bootstrap sampling to obtain different subsets to train the base learners, and AdaBoost \cite{freund1996experiments} uses adaptive sample weights (larger weights for harder samples) in generating a new base learner. 2) Feature-based strategies, which train each base learner on different subsets of features, and thus are scalable to high dimensionality. For example, each decision tree in Random Forest (RandomForest) \cite{breiman2001random,barandiaran1998random,amit1997shape} selects the feature to be split from a random subset of features, instead of all available features. Similarly, each decision tree in Extremely Randomized Trees (Extra-Trees) \cite{geurts2006extremely} splits nodes by drawing the cut-points completely randomly. 3) Parameter-based strategies. If the base learners are sensitive to their parameters, then setting different parameters can improve the diversity. For example, different hidden layer weights can be used to initialize diverse neural networks. Interestingly, these three categories of diversity enhancement strategies are complementary; so, they can be combined for better performance.

Boosting \cite{freund1996experiments,hastie2009multi,friedman2002stochastic}, the driving force of Gradient Boosting Machine (GBM) \cite{friedman2001greedy}, can be used to reduce the bias of an ensemble. It is an incremental learning process, in which a new base learner is built to compensate the error of previously generated learners. Each new base learner is added to the ensemble in a forward stage-wise manner. As the boosting algorithm iterates, base learners generated at later iterations tend to focus on the harder samples. Mason et al. \cite{mason2000} described boosting from the viewpoint of gradient descent and regarded boosting as a stage-wise learning scheme to optimize different objective functions iteratively. FilterBoost \cite{FilterBoost2007} is a variant of AdaBoost \cite{freund1996experiments}, which fits an additive logistic regression model and minimizes the negative log likelihood step-by-step. DeepBoosting \cite{DeepBoosting2014}  can use very deep decision trees as base learners. Popular implementations of GBM, e.g., XGBoost \cite{chen2016xgboost} and LightGBM \cite{ke2017lightgbm}, have been successfully used in many applications \cite{chen2015higgs,rakhlin2018deep,liu2017combining,walsh2019regularized}.  However, traditional boosting approaches \cite{friedman2001greedy,chen2016xgboost,ke2017lightgbm} often have many parameters and thus require cross-validation, which is unreliable on small datasets, and time-consuming on big data. It is desirable to develop an algorithm that has very few parameters to tune and is robust to them.

This paper proposes BoostForest, which integrates boosting and Bagging for both classification and regression. Our main contributions are:
\begin{enumerate}
\item We propose a novel decision tree model, BoostTree, that integrates GBM into a single decision tree, as shown in Fig.~\ref{fig:tree}. BoostTree uses the node function, e.g., RR, ELM, or SVR, to train a linear or nonlinear regression model at each node for regression or binary classification, or $J$ regression models for $J$-class classification, where $J$ is the number of classes. For a given input, BoostTree first sorts it down to a leaf, then computes the final prediction by summing up the outputs of all node models along the path from the root to that leaf. Similar to Extra-Trees, BoostTree increases the randomness (diversity) by drawing the cut-points randomly at node splitting.

\begin{figure}[htpb] \centering
\subfigure[]{\label{fig:tree}     \includegraphics[width=\linewidth,clip]{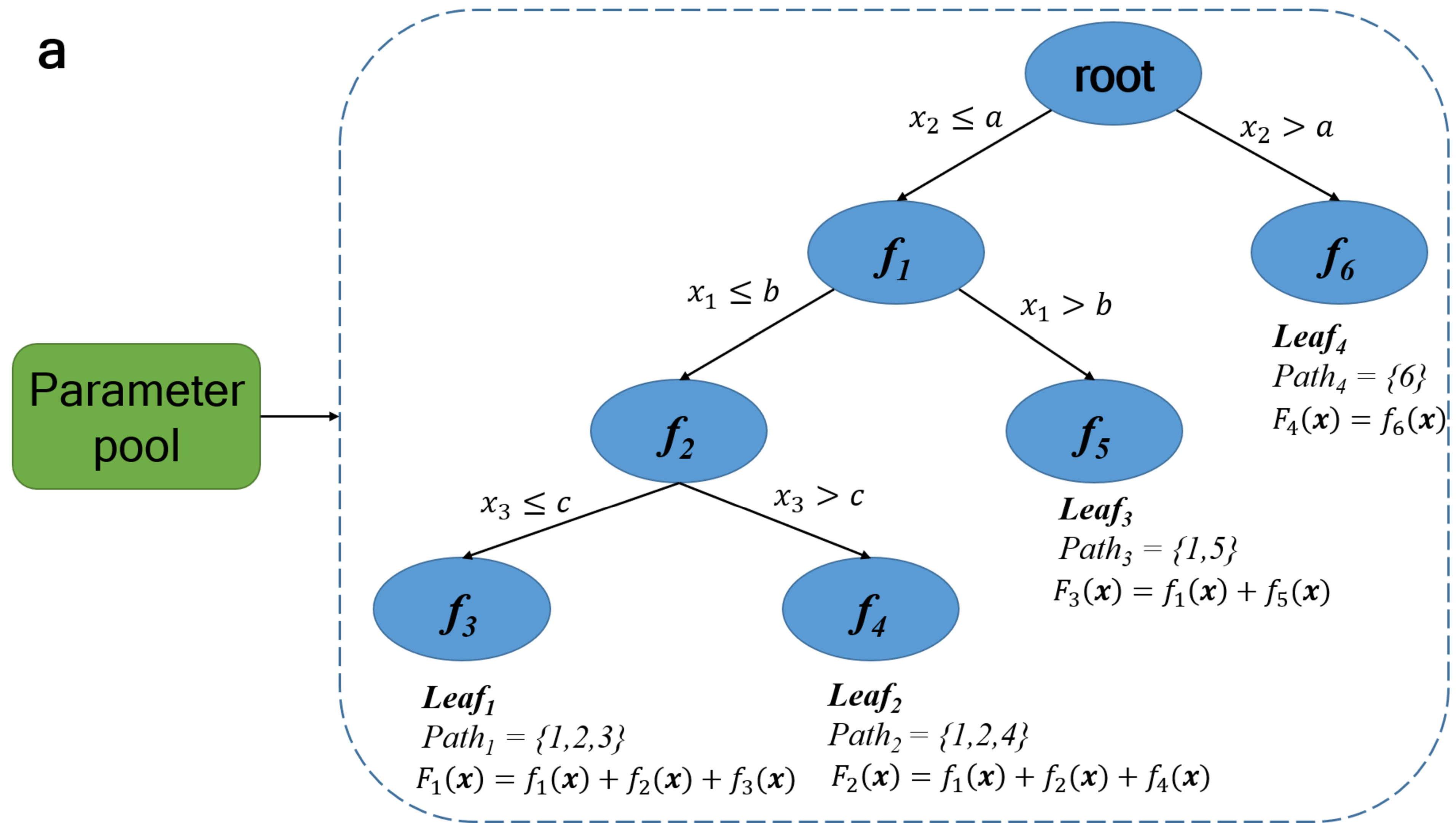}}
\subfigure[]{\label{fig:forest}     \includegraphics[width=\linewidth,clip]{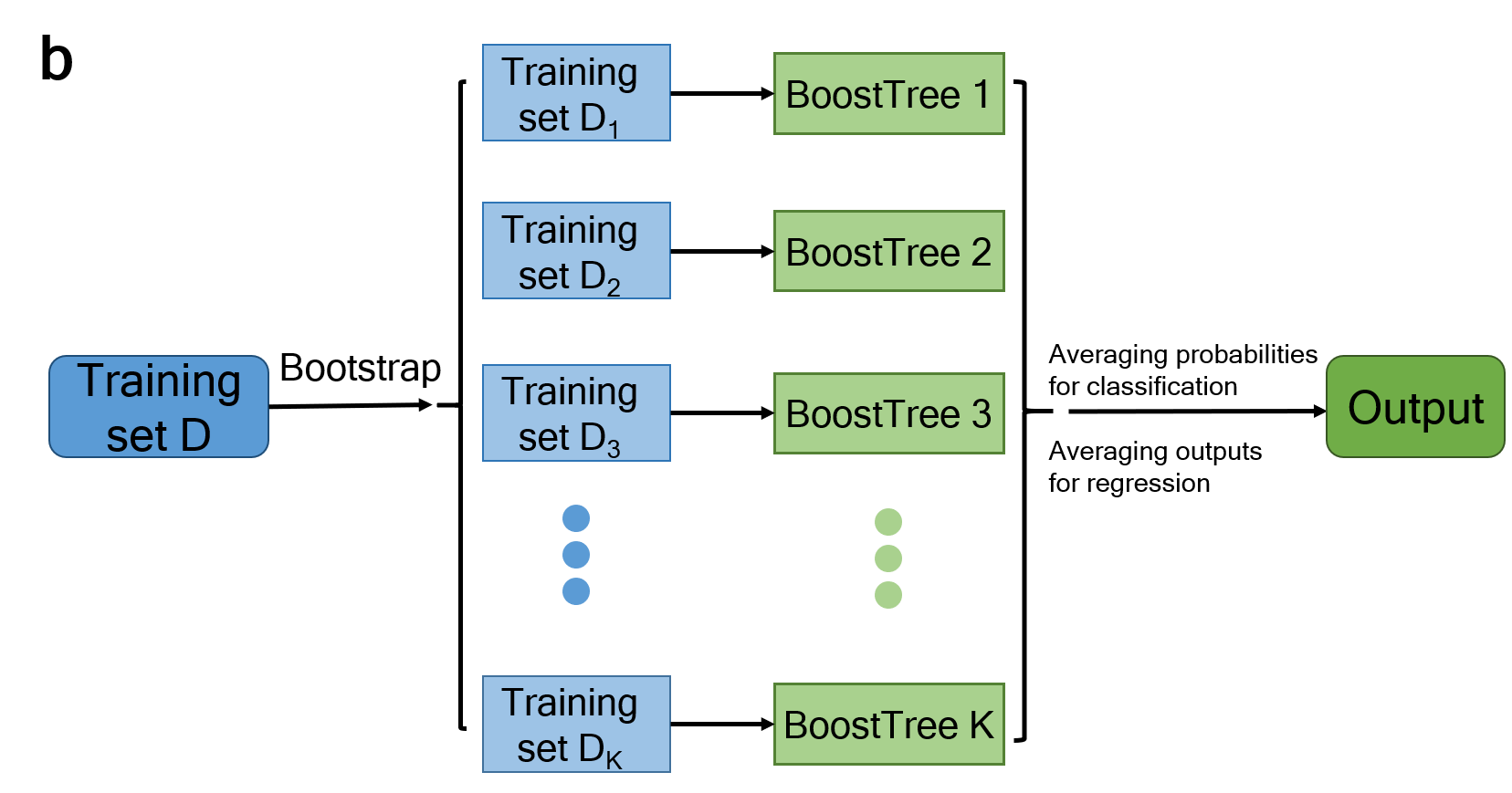}}
\caption{(a) BoostTree with four leaves. For a given input, BoostTree first sorts it down to a leaf $q(\bm{x})$, then computes the final prediction by summing up the outputs of all node models along the path (given by $\mathrm{Path}_{q(\bm{x})}$) from the root to the leaf. The parameters of BoostTree are randomly selected from a parameter pool. (b) BoostForest with $K$ BoostTrees. Bootstrap is used to obtain $K$ replicas of the training set.}
\end{figure}

\item We propose a novel parameter setting strategy, random parameter pool sampling, which makes BoostTree easier to tune its hyper-parameters than traditional approaches. In this strategy, the parameters of BoostTree are not specific values, but random samples from candidate sets stored in a parameter pool. Each time a root node is generated, BoostTree randomly selects its hyper-parameters from the parameter pool.
	
\item Using BoostTrees as base learners, we propose a novel ensemble learning approach, BoostForest, as shown in Fig.~\ref{fig:forest}. It first uses bootstrap to obtain multiple replicas of the original training set, and then trains a BoostTree on each replica. BoostForest uses the parameter pool sampling strategy to simplify its hyper-parameter tuning process, and outperforms several popular ensemble learning approaches. Moreover, it represents a very general ensemble learning framework, whose base learners can be any tree model, e.g., BoostTree, M5P, Extra-Tree (a single decision tree in Extra-Trees), or Logistic Model Tree (LMT) \cite{landwehr2005logistic}, or even a mixture of different models.
\end{enumerate}

The remainder of this paper is organized as follows: Section~\ref{sec:RelatedWork} introduces some related work. Section~\ref{sec:methods} describes the details of our proposed BoostTree and BoostForest. Section~\ref{sec:experiments} presents experimental results to demonstrate the superiority of BoostForest over several popular ensemble learning approaches on 35 classification and regression datasets. Finally, Section~\ref{sec:conclusion} draws conclusions and points out some future research directions.

\section{Related Work} \label{sec:RelatedWork}

GBM \cite{friedman2001greedy} generates an ensemble via an iterative process. It uses Newton's method to decompose the original classification or regression problem into multiple sub-regression problems, and solves them iteratively. Given a dataset $\mathcal{D}=\left\{\left(\bm{x}_n, y_n\right)\right\}_{n=1}^N$ with $N$ training samples, where $\bm{x}_n \in \mathbb{R}^{D\times1}$ and $D$ is the feature dimensionality, an ensemble $\phi$ generated by GBM uses $K$ base learners $\{f_k\}_{k=1}^K$ to predict the output:
\begin{align}
	\hat{y}_n=\phi\left(\bm{x}_n\right)=\sum_{k=1}^{K} f_k\left(\bm{x}_n\right). \label{eq:GBM1}
\end{align}

LogitBoost (Supplementary Algorithm~1) \cite{friedman2000additive} is a popular implementation of GBM to optimize logistic regression. In each iteration, LogitBoost first computes the pseudo-labels and weights using the Newton (for binary classification) or quasi-Newton (for multi-class classification) method, and then updates the ensemble by adding a new regression model (for binary classification) or $J$ new regression models [for $J$-class ($J>2$) classification], which are trained to fit the pseudo-labels by weighted least-squares regression. If the original problem is regression, then in each GBM iteration, the pseudo-label is the residual between the true value and the prediction, and all samples have the same weight.

LogitBoost further converts the GBM output into a probability:
\begin{align}
	\begin{cases}
	p(\bm{x}_n)=\mathrm{sigmoid}(\hat{y}_n),&J=2\\
	\bm{p}(\bm{x}_n)=[p^1(\bm{x}_n),...,p^J(\bm{x}_n)],&J>2
	\end{cases}, \label{eq:proba}
\end{align}
where
\begin{align}
	\mathrm{sigmoid}(\hat{y}_n)&=\frac{1}{1+e^{-\hat{y}_n}},\\
	p^j(\bm{x}_n)&=	\mathrm{softmax}^j(\hat{\bm{y}}_n)=\frac{e^{\hat{y}_n^j}}{\sum_{i=1}^Je^{\hat{y}_n^i}}, \label{eq:proba_j}
\end{align}
in which $\hat{\bm{y}}_n=[\hat{y}_n^1,\hat{y}_n^2,...,\hat{y}_n^J]^T\in \mathbb{R}^{J\times1}$ is the output vector for $J$-class ($J>2$) classification.

Malerba  et al. \cite{malerba2002trading} proposed Stepwise Model Tree Induction (SMOTI) to construct model trees gradually for regression. When the tree is growing, SMOTI adds more and more variables to the leaf model to approximate the target function. At each step, it adds either a splitting node or a regression node to the tree. A splitting node partitions the sample space, whereas a regression node removes the linear effect of the variables already included in its parent node and performs linear regression on one variable.

Different from GBM, LMT \cite{landwehr2005logistic} generates only one tree instead of multiple trees for predicting. LMT extends model trees from regression to classification, by integrating LogitBoost into the tree model. The final model at a leaf consists of all node models along the path from the root to that leaf.

Assume an LMT has trained a function $f_m$ for the $m$-th node. For an input $\bm{x}_n$, LMT first identifies $q(\bm{x}_n)$, the leaf node it belongs to, and then all $f_m$ along the path from the root to that leaf node are summed up to predict the output, i.e.,
\begin{align}
	\hat{y}_n=F\left(\bm{x}_n\right)=\sum_{m\in \mathrm{Path}_{q(\bm{x}_n)}} f_m\left(\bm{x}_n\right), \label{eq:LMT}
\end{align}
where $\mathrm{Path}_{q(\bm{x}_n)}$ is the collection of the node indices along the path from the root to the leaf node $q(\bm{x}_n)$. Similar to LogitBoost, LMT can use (\ref{eq:proba}) to output the probability.

SimpleLogistic \cite{landwehr2005logistic} (a variant of LogitBoost) incrementally refines the linear logistic model. In each iteration, instead of using all features to perform linear regression, SimpleLogistic uses only one feature to train the model. In this way, only relevant features are selected, and the risk of over-fitting is reduced.

Gradient Boosting with Piece-wise Linear Regression Trees (GBDT-PL) \cite{gbdtpl2019} is a variant of XGBoost, which combines multiple Piece-wise Linear Regression Trees (PL-Trees) to predict the output. Similar to LMT,  a PL-Tree fits a linear model at each node in an additive manner.

Webb  \cite{webb2000multiboosting} proposed MultiBoosting, which combines wagging (a variant of Bagging) and Adaboost \cite{freund1996experiments} to reduce both the bias and the variance. It uses AdaBoost to generate a sub-committee of decision trees. By wagging a set of sub-committees, MultiBoosting may achieve lower error than AdaBoost.

RandomForest \cite{breiman2001random} uses two techniques to improve the diversity of each tree, and hence reduces over-fitting: 1) Bagging, i.e., each tree is trained with a bootstrap replica drawn from the original training set; and, 2) feature sub-sampling, i.e., for each node of the tree, a subset of the features is randomly selected from the complete feature set, then an optimal feature is selected from the subset to split the node.

Extra-Trees \cite{geurts2006extremely} splits nodes in a more random way than RandomForest. It first randomly draws one cut-point for each feature, and then selects the cut-point with the largest splitting gain to split a node. Extra-Trees reduces over-fitting by trading the base learners' accuracy for diversity.

TAO \cite{TAO_NIPS}  first initializes the tree with a given depth by using CART \cite{breiman1984classification} or random parameters, and then circularly updates each decision or leaf node to optimize the objective function. It trains a binary classifier at each decision node to classify a sample into the left or right child node, and a constant label (TAO-c) or linear softmax classifier (TAO-l) at each leaf to predict the output. Its decision nodes are oblique nodes, which can split the samples more effectively than axis-aligned nodes. Bagged TAO-l (BaggedTAO-l) \cite{BaggedTAO} further uses Bagging to integrate multiple TAO-l trees for better performance.


\section{BoostTree and BoostForest} \label{sec:methods}

This section introduces the details of our proposed BoostTree and BoostForest. All source code is available at \url{https://github.com/zhaochangming/BoostForest}.

\subsection{Motivation}

Three problems need to be solved in model tree based ensemble learning:

\begin{enumerate}
	\item How to design a model tree that is more compatible with Bagging? Generally, as the model complexity increases, the bias of the model decreases, but the variance increases. Between the two popular ensemble learning strategies, Bagging is suitable for integrating complex base learners to reduce the variance, whereas boosting for integrating simple base learners to reduce the bias. MultiBoosting combines AdaBoost and wagging to further improve the performance of a decision tree. However, how to combine GBM and Bagging to improve the performance of model trees has not been studied. LMT is a stable model with low randomness, so simply combining multiple LMTs with Bagging may not outperform a single LMT.

	\item How to handle both classification and regression problems by using a single model tree structure? LMT applies LogitBoost to a decision tree to generate an accurate model tree; however, it cannot handle regression problems. GBDT-PL handles regression problems by using GBM to integrate multiple PL-Trees. However, performing boosting in a single tree is more efficient, and easier to be used in Bagging.

	\item How to generate a model tree with easy parameter tuning, so that any regression algorithm, e.g., RR, ELM and SVR, can be used as its node function?
\end{enumerate}

This section proposes BoostTree to solve the above three problems simultaneously. Moreover, multiple BoostTrees can be further integrated into a BoostForest for better performance.

\subsection{General Idea of BoostTree}

Our proposed BoostTree is inspired by LMT and GBM. Assume a BoostTree has $M$ nodes, excluding the root. Then, it trains a function $f_m$ for the $m$-th node, $m\in[1,M]$, and uses (\ref{eq:LMT}) to predict the output.

BoostTree minimizes the following regularized loss function:
\begin{align}
\mathcal{L}(F)& =\sum_{n=1}^N \ell\left(y_n, \hat{y}_n\right)+ \sum_{m=1}^M \lambda \Omega\left(f_m\right), \label{eq:BoostTree}
\end{align}
where $\lambda$ is the regularization coefficient. The second term above penalizes the complexity of BoostTree to reduce over-fitting.

Different loss functions $\ell$ can be used to cope with regression and classification problems, as will be introduced in Sections~\ref{sec:Regression}-\ref{sec:J-logiistic}. For the ease of optimization, we require $\ell$ to be convex and differentiable.

In general, the objective function in (\ref{eq:BoostTree}) cannot be optimized directly. Inspired by LMT and GBM, BoostTree minimizes  (\ref{eq:BoostTree}) in an additive manner. Assume a tree with $T$ $(T\geq2)$ leaves has been generated after $T-1$ iterations. Then, there are $M=2T-2$ nodes, excluding the root. We can rewrite (\ref{eq:BoostTree}) as:
\begin{align}
\mathcal{L}(F)& =\sum_{m=1}^T \mathrm{LeafLoss}_m+\sum_{m=1}^{2T-2} \lambda \Omega\left(f_m\right),\label{eq:BoostTree_}
\end{align}
where
\begin{align}
\mathrm{LeafLoss}_m& = \sum_{n\in I_m}\ell\left(y_n,\sum_{i \in \mathrm{Path}_m}f_i(\bm{x}_n)\right),\label{eq:LeafLoss}\\
I_m& = \{n|q(\bm{x}_n)=m \},
\end{align}
i.e., $I_m$ is the set of all training samples belonging to Leaf~$m$. $\mathrm{LeafLoss}_m$ measures the loss of Leaf~$m$. In each iteration, BoostTree uses a greedy learning scheme to add branches to the leaf with the highest loss.

Assume Node $m$ is the leaf node with the highest loss. After the split, $I_m$ is divided into two subsets: $I_L$ (of the left node) and $I_R$ (of the right node). Let $f_L$ and $f_R$ be the node models of the left and the right nodes trained separately using $I_L$ and $I_R$, respectively. Then, the reduction of the loss in (\ref{eq:BoostTree_}) is:
\begin{align}
	\delta_{\mathcal{L}} &=\sum_{n\in I_m}\ell(y_n,F_{m}(\bm{x}_n))-\sum_{n\in I_L}\ell(y_n,F_{m}(\bm{x}_n)+f_L(\bm{x}_n))\nonumber \\
	&-\sum_{n\in I_R}\ell(y_n,F_{m}(\bm{x}_n)+f_R(\bm{x}_n))-\lambda\Omega(f_L)-\lambda\Omega(f_R),\\
	&=\mathcal{C}-\mathcal{L}(f_L)-\mathcal{L}(f_R)
	\label{eq:GainSplit}
\end{align}
where
\begin{align}
	\mathcal{C} & = \sum_{n\in I_m}\ell(y_n,F_{m}(\bm{x}_n)),\nonumber \\
	F_{m}(\bm{x}_n) &= \sum_{i \in \mathrm{Path}_m}f_i(\bm{x}_n),\nonumber \\
	\mathcal{L}(f_L) & = \sum_{n\in I_L}\ell(y_n,F_{m}(\bm{x}_n)+f_L(\bm{x}_n))+\lambda\Omega(f_L), \nonumber \\
	\mathcal{L}(f_R) & = \sum_{n\in I_R}\ell(y_n,F_{m}(\bm{x}_n)+f_R(\bm{x}_n))+\lambda\Omega(f_R). \nonumber
\end{align}
Note that $\mathcal{C}$ is a constant, $F_{m}$ is the ensemble of the models along the path from the root to Node $m$, and $\mathcal{L}(f_L)$ and $\mathcal{L}(f_R)$ are the loss functions for the left and right child nodes, respectively.

(\ref{eq:GainSplit}) can be optimized by minimizing $\mathcal{L}(f_L)$ and $\mathcal{L}(f_R)$ separately. More specifically, BoostTree uses two steps to optimize (\ref{eq:GainSplit}):
\begin{enumerate}
	\item Split the node: BoostTree implements four criteria to select the cut-point:
\begin{enumerate}
\item XGBoost Splitting Criterion (XGB-SC), which uses gradient decent to reduce the loss.
\item Gini Splitting Criterion (Gini-SC), which tries to improve the purity of each leaf.
\item Mean Squared Error (MSE), which tries to reduce the MSE of each leaf.
\item C4.5 Splitting Criterion (C4.5-SC) \cite{c45}, which selects the cut-point with the largest ratio of information gain.
\end{enumerate}
Though all criteria are effective, XGB-SC achieved the best performance in our experiments, as will be demonstrated in Section~\ref{Sec:Ex_criterion}; hence, we make it the default option in BoostTree.
	
XGB-SC uses the Euclidean norm of the leaf score to constrain the tree complexity. Ignoring the parent node loss and inspired by Equation~(7) in XGBoost \cite{chen2016xgboost}, we can calculate the splitting gain as:
	\begin{align}
		\delta_{\mathrm{gain}} =&\frac{1}{2}\left[\frac{\left(\sum_{n \in I_{L}} g_{n}\right)^{2}}{\sum_{n \in I_{L}} h_{n}+\lambda}+\frac{\left(\sum_{n \in I_{R}} g_{n}\right)^{2}}{\sum_{n \in I_{R}} h_{n}+\lambda}\right], \label{eq:GainSplit2}
	\end{align}
	where
	\begin{align}
		g_{n}&=\frac{\partial \ell\left(y_n, F_{m}\left(\bm{x}_n\right)\right)}{\partial  F_{m}\left(\bm{x}_n\right)} \label{eq:g}\\
		h_{n}&=\frac{\partial^{2} \ell\left(y_n,  F_{m}\left(\bm{x}_n\right)\right)}{\partial  F_{m}\left(\bm{x}_n\right)^{2}} \label{eq:h}
	\end{align}
are the first and second order derivatives of the loss function w.r.t. $F_{m}\left(\bm{x}_n\right)$, respectively.
	
	Similar to Extra-trees for increasing the randomness, BoostTree first draws a random cut-point uniformly for each feature, and then determines the best cut-point of the current node according to the maximum $\delta_{\mathrm{gain}}$.

	\item Train the node model: BoostTree uses gradient boosting to decompose the original problem into multiple sub-regression problems, and uses a node function to solve a sub-regression problem in each node.  It first calculates the pseudo-labels and sample weights to generate a temporary training set, and then trains a regression model for regression or binary classification, or $J$ regression models for $J$-class ($J>2$) classification. For simplicity, RR is used as the default node function. More details on training the node model for different tasks are introduced in Sections~\ref{sec:Regression}-\ref{sec:J-logiistic}.
\end{enumerate}

For the ease of implementation, we can simply store all possible values of each parameter in a parameter pool, from which BoostTree randomly selects a combination of parameters, e.g., the minimum number of samples at a leaf $min\_samples\_leaf$, and the regularization coefficient $\lambda$.  Robustness of using different parameters to train a BoostTree in BoostForest will be studied in Section~\ref{Sec:robustness}.

Algorithm~\ref{alg:BoostTree} shows the pseudo-code of BoostTree using RR as the node function, where the subfunction \emph{FitModel} assumes different forms according to different learning tasks, as shown in Algorithms~\ref{Alg:LS}-\ref{Alg:LogisticMultiClass}.

\begin{algorithm}[htpb] \fontsize{10pt}{0.93\baselineskip}\selectfont
\caption{BoostTree using RR as the node function.}\label{alg:BoostTree}
\KwIn{$\mathrm{Data}=\{(\bm{x}_n,y_n)\}_{n=1}^N$, $N$ training samples, where $\bm{x}_n\in \mathbb{R}^{D\times1}$; $\mathrm{Pool}_\mathrm{MSL}$, candidate value pool of  the minimum number of samples at a leaf min\_samples\_leaf; $\mathrm{Pool}_{\lambda}$, candidate value pool of the $\ell_2$ regularization parameter $\lambda$; (optional) $\mathrm{MaxNumLeaf}$, the maximum number of leaves, default NULL.}
\KwOut{A BoostTree.}
Randomly select $\lambda$ and $\mathrm{MSL}$ from $\mathrm{Pool}_{\lambda}$ and $\mathrm{Pool}_\mathrm{MSL}$, respectively\;
Initialize $\mathrm{NumLeaf} = 1$, $\mathrm{LeafList} = \{\}$ and $f(\bm{x})=0$\;
$\mathrm{root\leftarrow \{data=Data,~ model=}$ $f$\}\;
\tcc{Initialize the root node.}
Add $\mathrm{root}$ to $\mathrm{LeafList}$\;
$\mathrm{BoostTree\leftarrow split(root)}$\;
\vspace*{2mm}
$\mathrm{split(node)}$\{\\
Let $m$ be the index of the current node\;
$\{(\bm{x}_n,y_n)\}_{n\in I_m}\leftarrow \mathrm{node.data}$\;
$F_{m}(\bm{x})\leftarrow \sum_{i \in \mathrm{Path}_m}\mathrm{node}_i.\mathrm{model}(\bm{x})$\;
\tcc{Identify the cut-point.}
Initialize $\delta_{\mathrm{gain}}^{\max}=0$ and $\mathrm{split}_\mathrm{flag}=\mathrm{false}$\;
\For{$d=1:D$}{
	Let $a_{\max}$ and $a_{\min}$ denote the maximal and minimal value of $ \{x_{n,d}|n\in I_m \}$, respectively\;
	Draw a random cut-point $s$ uniformly in $[a_{\min},~a_{\max}]$\;
	$I_L=\{n|x_{n,d}\le s, n\in I_m\}$\;
	$I_R=\{n|x_{n,d}>s, n\in I_m\}$\;
	\If{$|I_L|\geq \mathrm{MSL}$ $\mathrm{and}$ $|I_R|\geq \mathrm{MSL}$}
	{
		Calculate $\delta_{\mathrm{gain}}$ in  (\ref{eq:GainSplit2})\;
		\If{$\mathrm{split}_\mathrm{flag}==\mathrm{false}~\mathrm{or}~\delta_{\mathrm{gain}}>\delta_{\mathrm{gain}}^{\max}$}{
			$\delta_{\mathrm{gain}}^{\max}=\delta_{\mathrm{gain}}$, $\mathrm{split}_\mathrm{flag}=\mathrm{true}$, $I_L^*=I_L$, $I_R^*=I_R$\;
		}
	}
}
\If{$\mathrm{split}_\mathrm{flag}$ }
{
	\tcc{Train the node model.}
	$f_L^*=\mathrm{FitModel}(\{(\bm{x}_n,y_n)\}_{n\in I_L^*},F_m,\lambda)$\;
	$f_R^*=\mathrm{FitModel}(\{(\bm{x}_n,y_n)\}_{n\in I_R^*},F_m,\lambda)$\;
	\tcc{Generate the child nodes.}
	$\mathrm{node.leftChild}=\{\mathrm{data}=\{(\bm{x}_n,y_n)\}_{n\in I_L^*}$, $\mathrm{model}=f_L^*$  \}\;
	$\mathrm{node.rightChild=\{data}=\{(\bm{x}_n,y_n)\}_{n\in I_R^*}$, $\mathrm{model}=f_R^*$\}\;
	$\mathrm{NumLeaf=NumLeaf}+1$\;
	Add $\mathrm{node.leftChild}$ and $\mathrm{node.rightChild}$ to $\mathrm{LeafList}$\;
}
Pop $\mathrm{node}$ from $\mathrm{LeafList}$\;
\tcc{Split the leaf recursively.}
\If{$|\mathrm{LeafList}|>0$ $ \mathrm{and}$ $\mathrm{(MaxNumLeaf==NULL}$ $\mathrm{or}$ $\mathrm{NumLeaf}< \mathrm{MaxNumLeaf)}$}{
	Use (\ref{eq:LeafLoss}) to calculate the leaf loss of each $\mathrm{node}$ in $\mathrm{LeafList}$\;
	$\mathrm{split(node^*)}$, where $\mathrm{node^*}$ is the leaf node with the highest loss\;
}
\Return node\;
    \}
\end{algorithm}

\subsection{BoostTree for Regression} \label{sec:Regression}

For regression problems, we use
\begin{align}
	\ell(y_n,\hat{y}_n) =(y_n-\hat{y}_n)^2.
\end{align}

Assume Node $c$ is the left or right child node of Node $m$. According to GBM, the loss function for Node $c$ can be expressed as:
\begin{align}
	\mathcal{L}(f_c) & = \sum_{n\in I_c} \left\{y_n-[F_{m}(\bm{x}_n) + f_c(\bm{x}_n)]\right\}^2+  \lambda \|\bm{a}_c\|_2^2 \nonumber \\
	& = \sum_{n\in I_c} \left\{f_c(\bm{x}_n) - [y_n-F_{m}(\bm{x}_n)] \right\}^2+  \lambda \|\bm{a}_c\|_2^2 \nonumber \\
	&=\sum_{n\in I_c} \left[f_c(\bm{x}_n)-\tilde{y}_n\right]^2+  \lambda \|\bm{a}_c\|_2^2, \label{eq:Lfm}
\end{align}
where $f_c(\bm{x}_n)=\bm{a}_c^T\bm{x}_n+b_c$, $\tilde{y}_n=y_n-F_{m}(\bm{x}_n)$ is the pseudo-label, i.e., the residual between the true value and the prediction, $I_c$ is the set of all training samples belonging to Node $c$, $\bm{a}_c\in\mathbb{R}^{D\times1}$ is a vector of the regression coefficients, and $b_c$ is the intercept. Note that all samples have the same weight in regression.

MSE is sensitive to noise and outliers. To improve BoostTree's robustness, we save the lower (upper) bound $lb$ ($ub$) of each node's pseudo-label at the training stage, and clip the corresponding node model's output at the prediction stage:
\begin{align}
	\mathrm{Clip}\left(f_c(\bm{x}_n)\right)=\max\left(lb_c,\min\left(ub_c,f_c(\bm{x}_n)\right)\right), \label{eq:clip}
\end{align}
where $lb_c=\max\limits_{n\in I_c}\tilde{y}_n$ and $ub_c=\min\limits_{n\in I_c}\tilde{y}_n$.

When the tree is growing, the residuals of the nodes generated at later iterations approach zero, so $lb$ and $ub$ get closer. The effect of clipping will be demonstrated in Section~\ref{Sec:clipping}.

Algorithm~\ref{Alg:LS} shows the pseudo-code of BoostTree for regression.

\begin{algorithm}[htpb]
\caption{$\mathrm{FitModel}$ for regression.}\label{Alg:LS}
\KwIn{$\{(\bm{x}_n,y_n)\}_{n\in I_c}$, sample set of the current node\;
	\hspace*{10mm} $F_{m}$, ensemble of the models along the path from the root to the parent node of the current node\;
	\hspace*{10mm} $\lambda$, the $\ell_2$ regularization parameter.}
\KwOut{A regression model $f_c$ for the current node.}
\vspace*{2mm}
$\tilde{y}_n=y_n-F_{m}(\bm{x}_n)$, $n\in I_c$\;
Fit $f_c = \mathrm{RidgeRegression}(\{(\bm{x}_n,\tilde{y}_n)\}_{n\in I_c},\lambda)$ using RR with regularization parameter $\lambda$\;
Clip $f_c$ using (\ref{eq:clip}).
\end{algorithm}

\subsection{BoostTree for Binary Classification}\label{sec:2-logiistic}

In classification tasks, BoostTree is built using a LogitBoost-like algorithm, which iteratively updates the ensemble of the logistic linear models $F$ by adding a new regression model $f$ to it.

The cross-entropy loss is used in binary classification:
\begin{align}
\ell(y_n,\hat{y}_n) =&-y_n\log(\mathrm{sigmoid}(\hat{y}_n))\nonumber\\
& -(1-y_n)\log(1-\mathrm{sigmoid}(\hat{y}_n)),
\end{align}
where $\hat{y}_n = F_{m}(\bm{x}_n)$.

Assume Node $c$ is the left or right child node of Node $m$. According to LogitBoost (Supplementary Algorithm~1) \cite{friedman2000additive}, the second order Taylor expansion can be used to approximate the loss function for Node $c$:
\begin{align}
	\mathcal{L}(f_c)& =\sum_{n\in I_c}  \ell \left(y_n, F_{m}\left(\bm{x}_n\right) +f_c\left(\bm{x}_n\right)  \right) +  \lambda \|\bm{a}_c\|_2^2 \nonumber \\
	&\approx\sum_{n\in I_c} \left[  \ell\left(y_n, F_{m}\left(\bm{x}_n\right)  \right)+g_n f_c\left(\bm{x}_n\right) + \frac{1}{2}h_n {f_c\left(\bm{x}_n\right)}^2  \right]\nonumber \\ &+  \lambda \|\bm{a}_c\|_2^2 \nonumber \\
	&=\frac{1}{2}\sum_{n\in I_c}  h_n \left[f_c\left(\bm{x}_n\right) - (-\frac{g_n}{h_n}) \right]^2 +  \lambda \|\bm{a}_c\|_2^2 + \mathcal{C}, \label{c1}
\end{align}
where
\begin{align}
	\mathcal{C} &= \sum_{n\in I_c}   \ell\left(y_n, F_{m}\left(\bm{x}_n\right)  \right)- \sum_{n\in I_c} \frac{1}{2}\frac{(g_n)^2}{h_n} \nonumber
\end{align}
is a constant, and $g_n$ and $h_n$ are calculated by using (\ref{eq:g}) and (\ref{eq:h}), respectively. Note that $g_n$ and $h_n$ are irrelevant to $f_c$. Therefore, $f_c$ can be any regression model.

Then, we can construct the pseudo-label $\tilde{y}_n$ and the sample weight $w_n$, as in LogitBoost:
\begin{align}
	\tilde{y}_n&=-\frac{g_n}{h_n},\\
	w_n&=h_n,
\end{align}
where
\begin{align}
	g_n&=p(\bm{x}_n)-y_n,\\
	h_n&=p(\bm{x}_n)[1-p(\bm{x}_n)],
\end{align}
in which $p(\bm{x}_n)$ is the  estimated probability in (\ref{eq:proba}).

To improve BoostTree's robustness to outliers, we follow LogitBoost \cite{friedman2000additive} to limit the minimum weight to $2\epsilon$ ($\epsilon$ is the machine epsilon), and clip the value of the pseudo-label $\tilde{y}$ to:
\begin{align}
	\mathrm{Clip}\left(\tilde{y}\right)
	=\max\left(-y_{\max},\min\left(y_{\max},\tilde{y}\right)\right), \label{eq:clip_y}
\end{align}
where $y_{\max}\in [2,4]$ (according to LogitBoost \cite{friedman2000additive}). $y_{\max}=4$ was used in our experiments.

Finally, we can remove $\mathcal{C}$ in (\ref{c1}) to simplify the loss function for child Node $c$:
\begin{align}
	\mathcal{L}(f_c) =\frac{1}{2}\sum_{n\in I_c} w_n\left[f_c(\bm{x}_n)-\tilde{y}_n\right]^2+  \lambda \|\bm{a}_c\|_2^2.
\end{align}

Algorithm~3 shows the pseudo-code of BoostTree for binary classification.

\begin{algorithm}[htpb]
\caption{$\mathrm{FitModel}$ for binary classification.}\label{Alg:LogisticTwoClass}
\KwIn{$\{(\bm{x}_n,y_n)\}_{n\in I_c}$, sample set of the current node\;
\hspace*{10mm} $F_{m}$, ensemble of the models along the path from the root to the parent node of the current node\;
\hspace*{10mm} $\lambda$, the $\ell_2$ regularization parameter.}
\KwOut{A regression model $f_c$ for the current node.}
\vspace*{2mm}
$p(\bm{x}_n)=\mathrm{sigmoid}(F_{m}(\bm{x}_n))$, $n\in I_c$\;
$\displaystyle \tilde{y}_n=\frac{y_n-p(\bm{x}_n)}{p(\bm{x}_n)[1-p(\bm{x}_n)]}$, $n\in I_c$\;
$\tilde{y}_n=\mathrm{Clip}(\tilde{y}_n)$ in  (\ref{eq:clip_y}), $n\in I_c$\;
$w_n=p(\bm{x}_n)[1-p(\bm{x}_n)]$, $n\in I_c$\;
$\mathcal{D}'=\left\{\left(\bm{x}_n, \tilde{y}_n, w_n\right)|\ n\in I_c \right\}$\;
Fit $f_c = \mathrm{WeightedRidgeRegression}(\mathcal{D}',\lambda)$ using weighted RR on $\mathcal{D}'$ with regularization parameter $\lambda$.
\end{algorithm}

\subsection{BoostTree for $J$-Class ($J>2$) Classification} \label{sec:J-logiistic}

For $J$-class classification, we use
\begin{align}
\ell(\bm{y}_n,\hat{\bm{y}}_n) =-\sum_{j=1}^Jy^j_n\log\left(\mathrm{softmax}^j(\hat{\bm{y}}_n)\right),\label{eq:Entropy}
\end{align}
where $\bm{y}_n=[y_n^1,y_n^2,...,y_n^J]^T\in \mathbb{R}^{J\times1}$ is the one-hot encoding label vector, and $\hat{\bm{y}}_n=F_{m}(\bm{x}_n)=[\hat{y}_n^1,\hat{y}_n^2,...,\hat{y}_n^J]^T\in \mathbb{R}^{J\times1}$ is the output vector.

In each iteration, LogitBoost handles the $J$-class classification problem by decomposing it into $J$ regression problems, so $\bm{f}_c$ becomes a set of linear models $\{f_c^1$, $f_c^2$, $\cdots$, $f_c^J\}$, where $f_c^j$ calculates the output for Class~$j$.

Similarly, the second order Taylor expansion can be used to approximate the loss function for Node $c$:
\begin{align}
	\mathcal{L}(\bm{f}_c)& =\sum_{n\in I_c}   \ell \left(y_n, F_{m}\left(\bm{x}_n\right) +\bm{f}_c\left(\bm{x}_n\right)  \right) +  \lambda \sum_{j=1}^J\|\bm{a}_c^j\|_2^2 \nonumber \\
	&\approx\sum_{n\in I_c}   \ell\left(y_n, F_{m}\left(\bm{x}_n\right)  \right)+  \lambda \sum_{j=1}^J\|\bm{a}_c^j\|_2^2  \nonumber \\ &+ \sum_{j=1}^J\sum_{n\in I_c} \left[ g_n^j f_c^j\left(\bm{x}_n\right) + \frac{1}{2}h_n^j {f_c^j\left(\bm{x}_n\right)}^2 \right] \nonumber \\
	&=\frac{1}{2}\sum_{j=1}^J \sum_{n\in I_c}   h_n^j \left[f_c^j\left(\bm{x}_n\right) - (-\frac{g_n^j}{h_n^j}) \right]^2 +  \lambda \sum_{j=1}^J\|\bm{a}_c^j\|_2^2 \nonumber\\
	&+ \mathcal{C}, \label{c2}
\end{align}
where
\begin{align}
	\mathcal{C} = \sum_{n\in I_c}   \ell\left(y_n, F_{m}\left(\bm{x}_n\right)  \right)-\sum_{j=1}^J\sum_{n\in I_c} \frac{1}{2}\frac{(g_n^j)^2}{h_n^j}
\end{align}
is a constant, $\bm{a}_c^j$ is the coefficient vector of $f_c^j$, $g_n^j$ is the $j$-th element of $\bm{g}_n$, and $h_n^j$ is the $j$-th diagonal element of $\bm{h}_n$. $\bm{g}_n$ and $\bm{h}_n$ are again calculated by using (\ref{eq:g}) and (\ref{eq:h}), respectively.

Then, for the $j$-th class, we can calculate the pseudo-label $\tilde{y}^j_n$ and sample weight $w^j_n$, as in LogitBoost:
\begin{align}
	\tilde{y}_n^j&=-\frac{g_n^j}{h_n^j},\label{eq:pseudo2} \\
	w_n^j&=h_n^j, \label{eq:w2}
\end{align}
where
\begin{align}
	g^j_n&=p^j(\bm{x}_n)-y_n^j,\\
	h^j_n&=p^j(\bm{x}_n)[1-p^j(\bm{x}_n)],
\end{align}
in which $p^j(\bm{x}_n)$ is the estimated probability of Class~$j$ in (\ref{eq:proba_j}). To prevent too large pseudo-labels, we also use (\ref{eq:clip_y}) to clip $\tilde{y}^j_n$.

Finally, we can remove $\mathcal{C}$ in (\ref{c2}) to simplify the loss function for Node $c$:
\begin{align}
\mathcal{L}(\bm{f}_c) =\frac{1}{2}\sum_{j=1}^J  \sum_{n\in I_c} w_n^j \left[f_c^j\left(\bm{x}_n\right) -\tilde{y}_n^j \right]^2 + \sum_{j=1}^J \lambda \|\bm{a}_c^j\|_2^2.
\end{align}

Algorithm~\ref{Alg:LogisticMultiClass} shows the pseudo-code of BoostTree for $J$-class ($J>2$) classification.

\begin{algorithm}[htpb]
\caption{$\mathrm{FitModel}$ for $J$-class ($J>2$) classification.}\label{Alg:LogisticMultiClass}
\KwIn{$\{(\bm{x}_n,y_n)\}_{n\in I_c}$, sample set of the current node\;
\hspace*{10mm} $F_{m}$, ensemble of the models along the path from the root to the parent node of the current node\;
\hspace*{10mm} $\lambda$, the $\ell_2$ regularization parameter.}
\KwOut{The regression model set $\bm{f}_c$ for the current node.}
\vspace*{2mm}
Compute $p^j(\bm{x}_n)=\mathrm{softmax}^j(F_{m}(\bm{x}_n))$ , $n\in I_c$, $j=1,...,J$\;
\For{$j=1:J$}{
 $\displaystyle \tilde{y}_{n}^j=\frac{y_{n}^j-p^j(\bm{x}_n)}{p^j(\bm{x}_n)[1-p^j(\bm{x}_n)]}$, $n\in I_c$\;
$\tilde{y}_{n}^j=\mathrm{Clip}(\tilde{y}_{n}^j)$ in  (\ref{eq:clip_y}), $n\in I_c$\;
$w_{n}^j=p^j(\bm{x}_n)[1-p^j(\bm{x}_n)]$, $n\in I_c$\;
$\mathcal{D}'=\left\{\left(\bm{x}_n, \tilde{y}_{n}^j, w_{n}^j\right)|\ n\in I_c \right\}$\;
 Fit $f_c^j = \mathrm{WeightedRidgeRegression}(\mathcal{D}',\lambda)$ using weighted RR on $\mathcal{D}'$ with regularization parameter $\lambda$\;
}
\tcc{Center the outputs of the regression models, as in LogitBoost.}
$\displaystyle f_c^j(\bm{x})\leftarrow\frac{J-1}{J}\left[f_c^{j}(\bm{x})-\frac{1}{J}\sum_{i=1}^Jf_c^i(\bm{x})\right]$,\quad $j=1,...,J$\;
$\bm{f}_c=\{f_c^1,f_c^2,\dots,f_c^J\}$.
\end{algorithm}

\subsection{BoostForest}

BoostForest integrates multiple BoostTrees into a forest. It first uses bootstrap to generate $K$ replicas of the original training set, and then trains a BoostTree on each replica.

For regression, the outputs predicted by all $K$ BoostTrees are averaged as the final output:
\begin{align}
	\mathrm{BoostForest}(\bm{x})= 	\frac{1}{K}\sum_{k=1}^{K}\mathrm{BoostTree}_k (\bm{x}).
\end{align}

For classification, the probabilities predicted by all $K$ BoostTrees are averaged as the final probability:
\begin{align}
	&\mathrm{BoostForest}(\bm{x}) \nonumber \\
	=&\begin{cases}
		\frac{1}{K}\sum_{k=1}^{K}\mathrm{sigmoid}(\mathrm{BoostTree}_k (\bm{x})),&J=2\\
		\frac{1}{K}\sum_{k=1}^{K}\mathrm{softmax}(\mathrm{BoostTree}_k (\bm{x})),&J>2
	\end{cases}.
\end{align}

Algorithm~\ref{Alg:BoostForest} gives the pseudo-code of BoostForest.

\begin{algorithm}[htpb]
\caption{BoostForest training algorithm.}\label{Alg:BoostForest}
\KwIn{$\mathrm{Data}=\{(\bm{x}_n,y_n)\}_{n=1}^N$, $N$ training samples, where $\bm{x}_n\in\mathbb{R}^{d\times1}$\;
\hspace*{10mm} $\mathrm{n\_estimators}$, the number of BoostTrees\;
\hspace*{10mm} $\mathrm{Pool_{MSL}}$, parameter value pool of the minimum number of samples at a leaf\;
\hspace*{10mm} $\mathrm{Pool}_{\lambda}$, parameter value pool of the $\ell_2$ regularization parameter $\lambda$.}
\KwOut{A BoostForest.}
\vspace*{2mm}
$\mathrm{BoostForest}=\{\}$; \\
\For{$i=1: \mathrm{n\_estimators}$}{
    Bootstrap $\mathrm{Data}'$ from $\mathrm{Data}$\;
    Train $\mathrm{BoostTree}_i$ on $\mathrm{Data}'$ using Algorithm~\ref{alg:BoostTree}\;
    Add $\mathrm{BoostTree}_i$ to $\mathrm{BoostForest}$\;
}
\end{algorithm}

\subsection{Implementation Details}

Inspired by the row sampling trick in XGBoost and lightGBM, we use two tricks to improve the speed of BoostTree:
\begin{enumerate}
	\item Randomly select $BatchSize$ samples to identify the cut-point, if the number of samples belonging to the splitting node is larger than the batch size $BatchSize$.
	\item When the number of samples belonging to Node $m$ is larger than $BatchSize$, randomly select $BatchSize$ samples to train the node model, and approximate $\mathrm{LeafLoss}_m$ as:
	\begin{align}
		\mathrm{LeafLoss}_m& = \frac{|I_m|}{|I_\mathrm{batch}|}\sum_{n\in I_\mathrm{batch}}\ell\left(y_n,\sum_{i \in \mathrm{Path}_m}f_i(\bm{x}_n)\right),
	\end{align}
	where $I_\mathrm{batch}$ is the batch index set of $I_m$.
\end{enumerate}
$BatchSize=1,000$ was used in our experiments.

To ensure BoostTree's number of leave nodes is larger than or equal to two, BoostTree tries to split the root node until it is split successfully.

\subsection{Discussions}

To clarify the novelties of BoostForest, this subsection first discusses the differences and connections between it and some related approaches, e.g., LMT, GBDT-PL, MultiBoosting and TAO, and then summarizes approaches which motivate d BoostForest.

There are four main differences between LMT and BoostForest:
\begin{enumerate}
	\item LMT uses SimpleLogistic to train the node model, and only handles classification. BoostTree can use any regression algorithm to train the node model, and can handle both classification and regression.
	
	\item LMT computes the splitting gains on all values of the attributes to select the cut-point, whereas BoostTree draws the cut-points completely randomly, which increases its diversity and makes BoostTrees more suitable for Bagging.
	
	\item LMT prunes the tree to reduce over-fitting and uses only one tree to predict the output. However, a single learner can easily be affected by noise and outliers, leading to poor generalization. BoostForest combines multiple BoostTrees with Bagging to reduce over-fitting.
	
	\item LMT's asymptotic complexity is $O\left(d \cdot N \cdot \log N+N \cdot d^{2} \cdot Depth\right)$ (ignoring the pruning operation), where $Depth$ is the depth of the tree, $N$ is the sample size, and $D$ is the feature dimensionality. BoostTree's complexity is $O\left[d \cdot N+(N \cdot d^{2}+d^{3}) \cdot Depth\right]$ (using RR as the node function), where the first part is the cost of building an Extra-tree, and the second is the cost of building the node models. If down-sampling is performed before fitting the RR model, BoostTree's asymptotic complexity is less than $O\left[d \cdot N+(BatchSize \cdot d^{2}+d^{3}) \cdot M\right]$, where $M$ is the number of nodes.
\end{enumerate}

M5 and M5P train a linear model at each node during the training process. Their smoothing operations use all node models along the path from the root to a leaf to predict the output. M5 and M5P construct model trees for regression, whereas LMT constructs them for classification. LMT is the most relevant approach to BoostTree. They both integrate GBM into one single tree, train a base learner at each node to perform boosting, and combine all node models along the path from the root to the corresponding leaf to predict the output. BoostTree may be viewed as an upgraded LMT.

BoostTree improves LMT from three aspects:
\begin{enumerate}
	\item Model structure: BoostTree expands LMT's node models to non-linear models.
	\item Application scenario: BoostTree further extends LMT to regression problems.
	\item Training process: BoostTree optimizes the cut-point selection strategy by combining XGB-SC and Extra-trees' cut-point drawing strategy, and also the parameter setting strategy. These improvements make it easier to tune the hyper-parameters in BoostTree. Note that LMT can automatically perform cross-validation within each node to determine the number of training iterations for SimpleLogistic, and prune the tree to further reduce over-fitting. However, when the node models are replaced by other more complex models, the computational cost of cross-validation at all nodes may be very high. To alleviate this problem, BoostTree develops a random parameter pool sampling strategy and combines it with Bagging.
\end{enumerate}

There are four main differences between GBDT-PL and BoostForest:
\begin{enumerate}
	\item GBDT-PL trains a linear model at each node, and does not evaluate its performance on multi-class classification. BoostTree can use any regression algorithm to train the node model, and can handle binary and multi-class classifications, and regression.
	\item GBDT-PL splits the node and trains the node models simultaneously, whereas BoostTree first splits the node, and then trains the node models. When the node models are replaced by other complex models, it's very time-consuming for GBDT-PL to select the cut-point, because it needs to train the left and right node models at all candidate cut-points.
	\item GBDT-PL uses GBM to train multiple PL-Trees, whereas BoostTree integrate GBM into one single tree.
	\item GBDT-PL needs a validation set to select its hyper-parameters, which include the tree structure and node model parameters. Its computational cost increases with the number of hyper-parameter combinations. BoostTree optimizes its hyper-parameter setting strategy by integrating random parameter pool sampling and Bagging.
\end{enumerate}

When each node uses a linear model, BoostTree has a similar structure as a tree in GBDT-PL, because all linear node models along the path to the corresponding leaf can be combined into one single linear model. However, when each node uses a non-linear model, the node models cannot be easily combined, which makes the structures of BoostTree and GBDT-PL tree different.

There are three main differences between MultiBoosting and BoostForest:
\begin{enumerate}
	\item MultiBoosting uses traditional decision trees as the base learners, whereas BoostForest uses BoostTrees. Each BoostTree is trained by gradient boosting, so itself can be viewed as an ensemble model.
	\item MultiBoosting uses first AdaBoost to train multiple decision trees sequentially to reduce the bias, and then wagging (a variant of Bagging) to further reduce the variance. BoostForest first generates multiple BoostTrees independently, and then uses the idea of Random Forest to reduce their variance.
	\item MultiBoosting can only be used for classification, whereas BoostForest can handle both classification and regression.
\end{enumerate}

MultiBoosting and BoostForest both combine boosting and Bagging, but in different ways: MultiBoosting combines them serially, whereas BoostForest combines them in parallel.

TAO is a general and efficient optimization algorithm, which can train many types of decision trees, e.g., both TAO and BoostTree can train the base learners in a tree ensemble. However, there are three main differences between TAO and BoostTree:
\begin{enumerate}
	\item TAO trains a binary classifier at each decision node to classify a sample into the left or right child node. Its decision nodes are oblique nodes, whereas BoostTree's decision nodes are axis-aligned nodes.
	\item TAO uses only the leaf model to predict the output, whereas BoostTree combines all node models along the path from the root to the corresponding leaf to predict the output.
	\item TAO circularly updates each decision or leaf nodes to optimize the objective function, whereas BoostTree optimizes it by using a greedy learning scheme to add new branches to the leaf.
\end{enumerate}

Table~\ref{tab:motivation} summarizes the approaches that motivated BoostForest. Essentially, BoostForest integrates five existing strategies and develops two new empirical ones.

\begin{table}[!h]
	\caption{Approaches that motivated BoostForest.} \label{tab:motivation}
	\begin{tabular}{@{}c|c@{}}
		\toprule
		BoostForest choice                         & Motivated by            \\ \midrule
		Tree structure                             & LMT \cite{landwehr2005logistic}                     \\ \midrule
		Splitting criteria                         & XGBoost \cite{chen2016xgboost}, CART \cite{breiman1984classification} and C4.5 \cite{c45}  \\ \midrule
		\makecell[c]{Cut-point drawing}             & Extra-trees \cite{geurts2006extremely}             \\ \midrule
			 	
		\makecell[c]{Node model training}        &  \makecell[c]{LMT \cite{landwehr2005logistic}, GBM \cite{friedman2001greedy}, LogitBoost  \cite{friedman2000additive} and \\ stochastic gradient boosting \cite{friedman2002stochastic}} \\ \midrule
		\makecell[c]{Tree ensemble generation}    & RandomForest \cite{breiman2001random}            \\ \midrule
		\makecell[c]{Random parameter\\ pool sampling}  & Empirical               \\ \midrule
		\makecell[c]{Clipping operation\\ in regression}           & Empirical               \\ \bottomrule
	\end{tabular}
\end{table}

\section{Experimental Results}\label{sec:experiments}

Extensive experiments were carried out to verify the effectiveness of BoostForest in both classification and regression. The following seven questions were examined:
\begin{enumerate}
\item What is the generalization performance of BoostForest, compared with several popular ensemble learning approaches, e.g., RandomForest \cite{breiman2001random}, Extra-Trees \cite{geurts2006extremely}, XGBoost \cite{chen2016xgboost}, LightGBM \cite{ke2017lightgbm}, MultiBoosting \cite{webb2000multiboosting}, GBDT-PL \cite{gbdtpl2019}, FilterBoost \cite{FilterBoost2007} and DeepBoosting \cite{DeepBoosting2014}?
\item How fast does BoostForest converge, as the number of base learners increases?
\item How does the base learner model complexity affect the generalization performance of BoostForest?
\item Can BoostForest handle datasets with a large number of samples or features?
\item Can our proposed approach for constructing BoostForest, i.e., data replica by bootstrapping and random parameter selection from the parameter pool, also be used to integrate other tree models, e.g., Extra-Tree \cite{geurts2006extremely}, M5P \cite{wang1997induction}, and LMT \cite{landwehr2005logistic}?
\item How does the performance of BoostForest change when different node functions, e.g., RR, ELM or SVR, are used in BoostTrees?
\item How robust is BoostForest to its hyper-parameters and the node splitting criterion?
\end{enumerate}

\subsection{Datasets}

We performed experiments on 37 real-world datasets\footnote{http://archive.ics.uci.edu/ml/datasets.php; http://www.cad.zju.edu.cn/home/dengcai/Data/data.html; http://yann.lecun.com/exdb/mnist/ } (20 for classification and 17 for regression), summarized in Supplementary Table~1. They covered a wide range of feature dimensionalities (between 4 and 1,024) and sample sizes (between 103 and 11,000,000). For each dataset, categorical features were converted to numerical ones by one-hot encoding. Unless stated otherwise, each feature's distribution was scaled to a standard normal distribution, and the labels in the regression datasets were $z$-normalized.

\subsection{Algorithms}

Supplementary Table~2 shows the hyper-parameters of the 11 baselines and BoostForest used in our experiments. More details about the baseline parameters can be found online\footnote{https://scikit-learn.org;\\ https://xgboost.readthedocs.io/en/latest/parameter.html; https://lightgbm.readthedocs.io/en/latest/Parameters.html; https://github.com/GBDT-PL/GBDT-PL; https://github.com/dmarcous/deepboost; https://www.cs.waikato.ac.nz/ml/weka/}. Unless stated otherwise, BoostForest used default parameter values.

\subsection{Experimental Setting}

We performed experiments on 30 small to medium sized datasets (15 for classification and 15 for regression) in Sections~\ref{sec:question1}-\ref{sec:question3} and \ref{sec:question7}, 35 datasets (30 small to medium sized ones and five large sized ones) in Section~\ref{sec:question4}, 32 datasets (30 small to medium sized ones and two high dimensional ones) in Section~\ref{sec:question5}, and 17 datasets (15 small to medium sized ones and two large sized ones) in Section  \ref{sec:question6}.

All algorithms were repeated 10 times on each dataset (except MNIST and HIGGS). For each experiment, we first randomly divided the data into 60\% training, 20\% validation and 20\% test. For MNIST or HIGGS, we randomly divided the data into 80\% training, 20\% validation, and used the original test data for testing. Next, we used the validation set to select the best parameters. Then, the training set and validation set were combined to train a model with these parameters. Finally, we verified its performance on the test set. For RandomForest and Extra-Trees, we used the out-of-bag error to select their parameters.

We used the classification accuracy and the root mean squared error (RMSE) as the main performance measure for classification and regression, respectively. We also computed the average rank and training time (including the process of parameter selection) for each algorithm on each dataset. For $K$ algorithms, the best one has rank $1$, and the worst has rank $K$. We used seconds as the unit of time in our experiments and 50 processing cores to run the experiments on a server with two Intel(R) Xeon(R) Platinum 8276 CPUs. Additionally, we used bytes as the unit of model size in our experiments.

To validate if BoostForest significantly outperformed the baselines ($\alpha=0.05$), we first calculated the $p$-values using the standard $t$-test, and then performed Benjamini Hochberg False Discovery Rate (BH-FDR) correction \cite{benjamini1995controlling} to adjust them. The statistically significant ones are marked by $\bullet$.

\subsection{Generalization Performance of BoostForest} \label{sec:question1}

First, we compared the generalization performance of BoostForest with eight baselines. The results are shown in Tables~\ref{tab:Ex1} and \ref{tab:Ex1_binary}.

\begin{table*} \centering  \setlength{\tabcolsep}{2mm} 
  \caption{Performances of the seven ensemble learning approaches on the 30 datasets. The best performance is marked in bold. $\bullet$ indicates a statistically significant win for BoostForest. } \label{tab:Ex1}
  \scalebox{1}{
  	\begin{tabular}{c|llllll}
  		\toprule
  		\multicolumn{1}{c|}{Classification Dataset} & RandomForest & Extra-Trees  & XGBoost & LightGBM & MultiBoosting & BoostForest\\
  		\midrule
  		&\multicolumn{6}{c}{Mean and standard deviation (in parentheses) of the classification accuracy } \\
  		\midrule
  		SON              & 0.8500 (0.0500)            & 0.8595 (0.0607)            & $\textbf{0.8619}$ (0.0551) & 0.8405 (0.0648)          & 0.8429 (0.0641)            & 0.8500 (0.0574)            \\
  		SEE              & 0.9143 (0.0265)$\bullet$   & 0.9310 (0.0225)$\bullet$   & 0.9238 (0.0233)$\bullet$   & 0.9238 (0.0333)$\bullet$ & 0.9500 (0.0225)$\bullet$   & $\textbf{0.9667}$ (0.0117) \\
  		QB               & $\textbf{1.0000}$ (0.0000) & 0.9980 (0.0060)            & 0.9880 (0.0098)$\bullet$   & 0.9880 (0.0098)$\bullet$ & 0.9940 (0.0092)            & \textbf{1.0000} (0.0000)            \\
  		VC2              & 0.8097 (0.0455)$\bullet$   & 0.8355 (0.0493)            & 0.8242 (0.0429)            & 0.8290 (0.0416)          & 0.8226 (0.0346)$\bullet$   & $\textbf{0.8500}$ (0.0361) \\
  		VC3              & 0.8129 (0.0383)$\bullet$   & 0.8065 (0.0279)$\bullet$   & 0.8226 (0.0495)            & 0.8161 (0.0480)          & 0.8032 (0.0426)$\bullet$   & $\textbf{0.8371}$ (0.0255) \\
  		MV1              & 0.8812 (0.0310)            & 0.9010 (0.0243)            & 0.9042 (0.0298)            & 0.8969 (0.0341)          & $\textbf{0.9094}$ (0.0255) & 0.8917 (0.0284)            \\
  		BCD              & 0.9605 (0.0153)$\bullet$   & 0.9623 (0.0192)$\bullet$   & 0.9632 (0.0238)$\bullet$   & 0.9640 (0.0173)$\bullet$ & 0.9781 (0.0137)            & $\textbf{0.9798}$ (0.0136) \\
  		ILP              & 0.7051 (0.0390)$\bullet$   & $\textbf{0.7333}$ (0.0278) & 0.6872 (0.0349)$\bullet$   & 0.7128 (0.0297)          & 0.7000 (0.0394)$\bullet$   & 0.7274 (0.0347)            \\
  		BD               & 0.7827 (0.0272)            & 0.7813 (0.0139)            & $\textbf{0.7833}$ (0.0201) & 0.7773 (0.0205)          & 0.7660 (0.0282)            & 0.7807 (0.0156)            \\
  		PID              & 0.7662 (0.0186)            & $\textbf{0.7682}$ (0.0205) & 0.7604 (0.0305)            & 0.7617 (0.0317)          & 0.7519 (0.0223)            & 0.7682 (0.0197)            \\
  		VS               & 0.7400 (0.0162)$\bullet$   & 0.7376 (0.0241)$\bullet$   & 0.7565 (0.0286)$\bullet$   & 0.7594 (0.0247)$\bullet$ & 0.7529 (0.0207)$\bullet$   & $\textbf{0.8429}$ (0.0215) \\
  		QSAR             & 0.8801 (0.0221)            & 0.8754 (0.0188)$\bullet$   & 0.8673 (0.0246)$\bullet$   & 0.8758 (0.0229)          & 0.8735 (0.0249)$\bullet$   & $\textbf{0.8934}$ (0.0225) \\
  		DRD              & 0.6654 (0.0273)$\bullet$   & 0.6814 (0.0252)$\bullet$   & 0.6745 (0.0282)$\bullet$   & 0.6779 (0.0233)$\bullet$ & 0.6870 (0.0227)$\bullet$   & $\textbf{0.7442}$ (0.0192) \\
  		BA               & 0.9942 (0.0033)$\bullet$   & 0.9993 (0.0015)            & 0.9953 (0.0040)$\bullet$   & 0.9949 (0.0044)$\bullet$ & 0.9985 (0.0018)$\bullet$   & $\textbf{1.0000}$ (0.0000) \\
  		WDG              & 0.8579 (0.0075)$\bullet$   & 0.8648 (0.0063)$\bullet$   & 0.8609 (0.0086)$\bullet$   & 0.8611 (0.0087)$\bullet$ & 0.8572 (0.0077)$\bullet$   & $\textbf{0.8716}$ (0.0102) \\
  		\midrule
  		Average accuracy & 0.8413 (0.0079)            & 0.8490 (0.0077)            & 0.8449 (0.0122)            & 0.8453 (0.0086)          & 0.8458 (0.0097)            & $\textbf{0.8669}$ (0.0064) \\
  		Average rank     & 4.4667                     & 3.0667                     & 3.8667                     & 3.8000                   & 4.1333                     & $\textbf{1.6667}$          \\
  		Average time                      & 10.6116 (0.1849)                & 11.0003 (0.0768)                & 17.1842 (0.1357)                     & 7.5803 (0.5097)                       & \textbf{3.8320} (0.2296)                        & 4.7953 (0.0896)            \\
  		Average model size                      & $3.5578\times10^5$                 &  $6.4825\times10^5$              &  $1.0264\times10^5$                          & $\bm{5.9302\times10^4}$                   & $1.3214\times10^5$                       & $5.7421\times10^{6}$                 \\
  		\midrule\midrule
  		\multicolumn{1}{c|}{Regression Dataset} & RandomForest & Extra-Trees  & XGBoost & LightGBM & GBDT-PL & BoostForest\\
  		\midrule
  		&\multicolumn{6}{c}{Mean and standard deviation (in parentheses) of the regression RMSE} \\
  		\midrule
  		CS           & 0.4041 (0.0684)$\bullet$   & 0.3506 (0.0629)$\bullet$ & 0.3500 (0.0994)$\bullet$   & 0.3913 (0.0597)$\bullet$ & 0.3420 (0.1044)$\bullet$   & $\textbf{0.2284}$ (0.0331) \\
  		CF           & 0.7748 (0.1174)$\bullet$   & 0.7461 (0.0940)$\bullet$ & 0.8291 (0.1136)$\bullet$   & 0.7847 (0.0895)$\bullet$ & 0.8493 (0.1406)$\bullet$   & $\textbf{0.6991}$ (0.0721) \\
  		AMPG         & 0.3621 (0.0587)$\bullet$   & 0.3520 (0.0694)          & 0.3734 (0.0533)$\bullet$   & 0.3662 (0.0600)          & 0.3733 (0.0585)$\bullet$   & $\textbf{0.3422}$ (0.0706) \\
  		REV          & $\textbf{0.5087}$ (0.0960) & 0.5093 (0.0938)          & 0.5485 (0.0910)            & 0.5261 (0.0880)          & 0.5628 (0.0904)$\bullet$   & 0.5161 (0.1005)            \\
  		NO           & $\textbf{0.6375}$ (0.0628) & 0.6527 (0.0777)          & 0.6412 (0.0664)            & 0.6499 (0.0906)          & 0.6436 (0.0727)            & 0.6422 (0.0750)            \\
  		PM           & 0.8051 (0.0478)$\bullet$   & 0.8024 (0.0457)$\bullet$ & 0.7654 (0.0598)            & 0.7493 (0.0522)          & $\textbf{0.7422}$ (0.0604) & 0.7786 (0.0427)            \\
  		BH           & 0.4009 (0.0828)$\bullet$   & 0.3895 (0.0742)$\bullet$ & 0.3962 (0.0844)$\bullet$   & 0.4002 (0.0821)$\bullet$ & 0.4119 (0.0968)$\bullet$   & $\textbf{0.3593}$ (0.0705) \\
  		CPS          & 0.8354 (0.0646)            & 0.8295 (0.0768)          & $\textbf{0.8273}$ (0.0606) & 0.8346 (0.0665)          & 0.8489 (0.0598)            & 0.8682 (0.0745)            \\
  		CCS          & 0.2808 (0.0167)$\bullet$   & 0.2763 (0.0191)$\bullet$ & 0.2472 (0.0256)            & 0.2390 (0.0209)          & $\textbf{0.2367}$ (0.0225) & 0.2529 (0.0192)            \\
  		ASN          & 0.2486 (0.0164)            & 0.2333 (0.0194)          & 0.2239 (0.0164)            & 0.2283 (0.0124)          & $\textbf{0.2177}$ (0.0318) & 0.2478 (0.0193)            \\
  		ADS          & 0.6552 (0.0286)            & 0.6598 (0.0300)          & 0.6564 (0.0286)            & 0.6586 (0.0267)          & 0.6494 (0.0284)            & $\textbf{0.6493}$ (0.0187) \\
  		WQW          & 0.6972 (0.0379)$\bullet$   & 0.6924 (0.0365)          & 0.7108 (0.0337)$\bullet$   & 0.7073 (0.0352)$\bullet$ & 0.6918 (0.0320)            & $\textbf{0.6874}$ (0.0324) \\
  		AQ           & 0.0032 (0.0030)$\bullet$   & 0.0038 (0.0037)$\bullet$ & 0.0030 (0.0030)            & 0.0074 (0.0031)$\bullet$ & 0.0031 (0.0011)$\bullet$   & $\textbf{0.0016}$ (0.0013) \\
  		CCPP         & 0.1933 (0.0051)$\bullet$   & 0.1972 (0.0053)$\bullet$ & 0.1792 (0.0053)            & 0.1784 (0.0061)          & $\textbf{0.1773}$ (0.0051) & 0.1902 (0.0062)            \\
  		EGSS         & 0.3188 (0.0051)$\bullet$   & 0.3036 (0.0040)$\bullet$ & 0.2133 (0.0036)            & 0.2073 (0.0042)          & $\textbf{0.1894}$ (0.0042) & 0.2195 (0.0040)            \\
  		\midrule
  		Average RMSE & 0.4750 (0.0122)            & 0.4666 (0.0114)          & 0.4643 (0.0137)            & 0.4619 (0.0134)          & 0.4626 (0.0216)            & $\textbf{0.4455}$ (0.0109) \\
  		Average rank & 4.2000                     & 3.9333                   & 3.4000                     & 3.7333                   & 3.0667                     & $\textbf{2.6667}$            \\
  		Average time                      & 10.1125 (0.2117)                & 10.9757 (0.1039)                & 19.9508 (0.1587)                     & 12.1506 (0.4130)                      & 75.4997 (0.2494)                        & \textbf{4.8862} (0.1055)            \\
  		Average model size                      & $4.0171\times10^{6}$                 &  $5.1793\times10^{6}$              &  $5.0651\times10^{5}$                          & $\bm{3.8676\times10^{5}}$                    & $1.1150\times10^{6}$                       & $9.1213\times10^{6}$                 \\
  		\bottomrule
  \end{tabular} }
\end{table*}%

\begin{table}\centering \setlength{\tabcolsep}{1mm}
	\caption{Performances of the four ensemble learning approaches on the 11 binary classification datasets. The best performance is marked in bold. $\bullet$ indicates a statistically significant win for BoostForest. } \label{tab:Ex1_binary}
	\scalebox{0.8}{
	\begin{tabular}{c|llll}
		\toprule
		\makecell[c]{Classification \\ Dataset}                        & GBDT-PL             & DeepBoosting             & FilterBoost             & BoostForest                  \\ \midrule
		&\multicolumn{4}{c}{Mean and standard deviation (in parentheses)}
		\\
		&\multicolumn{4}{c}{of the classification accuracy} \\
		\midrule
		SON              & 0.8143 (0.0810)            & 0.8071 (0.1024)          & 0.8452 (0.0586)            & $\textbf{0.8500}$ (0.0574) \\
		QB               & 0.9980 (0.0060)            & 0.9880 (0.0098)$\bullet$ & 0.9880 (0.0098)$\bullet$   & $\textbf{1.0000}$ (0.0000) \\
		VC2              & 0.7968 (0.0383)$\bullet$   & 0.8097 (0.0594)$\bullet$ & 0.8048 (0.0435)$\bullet$   & $\textbf{0.8500}$ (0.0361) \\
		MV1              & 0.8958 (0.0349)            & 0.8583 (0.0393)          & $\textbf{0.8990}$ (0.0361) & 0.8917 (0.0284)            \\
		BCD              & 0.9632 (0.0348)            & 0.9456 (0.0321)$\bullet$ & 0.9693 (0.0119)$\bullet$   & $\textbf{0.9798}$ (0.0136) \\
		ILP              & 0.6761 (0.0339)$\bullet$   & 0.6897 (0.0342)$\bullet$ & 0.6932 (0.0252)$\bullet$   & $\textbf{0.7274}$ (0.0347) \\
		BD               & $\textbf{0.7813}$ (0.0217) & 0.7773 (0.0225)          & 0.7780 (0.0215)            & 0.7807 (0.0156)            \\
		PID              & 0.7591 (0.0272)            & 0.7442 (0.0250)$\bullet$ & 0.7571 (0.0210)            & $\textbf{0.7682}$ (0.0197) \\
		QSAR             & 0.8664 (0.0358)$\bullet$   & 0.8735 (0.0220)$\bullet$ & 0.8640 (0.0222)$\bullet$   & $\textbf{0.8934}$ (0.0225) \\
		DRD              & 0.7316 (0.0361)            & 0.6775 (0.0228)$\bullet$ & 0.6710 (0.0301)$\bullet$   & $\textbf{0.7442}$ (0.0192) \\
		BA               & 0.9967 (0.0025)$\bullet$   & 0.9895 (0.0084)$\bullet$ & 0.9971 (0.0022)$\bullet$   & $\textbf{1.0000}$ (0.0000) \\ \midrule
	Average accuracy & 0.8436 (0.0115)            & 0.8328 (0.0137)          & 0.8424 (0.0076)            & $\textbf{0.8623}$ (0.0076) \\
		Average rank     & 2.6364                     & 3.4545                   & 2.6364                     & $\textbf{1.2727}$          \\
		Average time                      & 96.0279 (0.6398)                        & 11.2469 (0.1279)                     & \textbf{1.2163} (0.1139)                & 2.9423 (0.0529)                     \\
		Average model size                      & $1.0386\times10^6$                 & $\bm{1.9835\times10^3}$               &  $7.0042\times10^3$                          & $2.7449\times10^{6}$         \\
		\bottomrule
	\end{tabular}}
\end{table}

Table~\ref{tab:Ex1} shows that BoostForest achieved the best generalization performance on 17 out of the 30 datasets, the fastest average time in regression, the second fastest average time in classification, and the best average performance, standard deviation and rank in both classification and regression.

Table~\ref{tab:Ex1_binary} shows that BoostForest achieved the best generalization performance on nine out of the 11 datasets, the best average performance, standard deviation and rank, and the second fastest average time.

On average, BoostForest achieved the best classification accuracy or regression RMSE. It significantly outperformed RandomForest on 19 datasets, Extra-Trees on 15 datasets, XGBoost on 14 datasets, LightGBM on 12 datasets, MultiBoosting on nine datasets, GBDT-PL on 10 datasets, DeepBoosting on eight datasets, and FilterBoost on seven datasets.

It is also important to note that BoostForest used the default parameter settings on all datasets, and did not use a validation set to select the best parameters. So, it is very convenient to use. Additionally, each BoostTree is trained with a different training set, so BoostForest can be easily parallelized to further speed it up.

One limitation of BoostForest is that its average model size was the largest, so it may not be easily deployed to low storage devices. One of our future research directions is to reduce its model size.

\subsection{Generalization Performance w.r.t. the Number of Base Learners} \label{sec:question2}

BoostForest needs to specify the number of BoostTrees in it. It is important to study how its performance changes with this number.

On each dataset, we gradually increased the number of base learners from three to 250, and tuned other parameters of the four baselines by using the validation set (for XGBoost and LightGBM) or the out-of-bag error (for RandomForest and Extra-Trees).

Fig.~\ref{fig:C2} shows the accuracies of the five algorithms on the last four classification datasets. Complete results on all 15 classification datasets are shown in Supplementary Fig.~1. Generally, as the number of base learners increased, the performances of all ensemble learning approaches quickly converged. BoostForest achieved the highest classification accuracy on 11 of the 15 datasets, and the second highest on another three (MV1, BD, and PID).

Fig.~\ref{fig:R2} shows the RMSEs of the five algorithms on the last four regression datasets. Complete results on all 15 regression datasets are shown in Supplementary Fig.~2. Again, as the number of base learners increased, generally the performances of all algorithms quickly converged. BoostForest achieved the smallest RMSE on seven of the 15 datasets.

In summary, BoostForest has very good generalization performance w.r.t. the number of base learners, and it generally converges faster than the four baselines (RandomForest, Extra-Trees, XGBoost, and LightGBM).

\begin{figure*} \centering
	\subfigure[]{\label{fig:C2}     \includegraphics[width=.95\linewidth,clip]{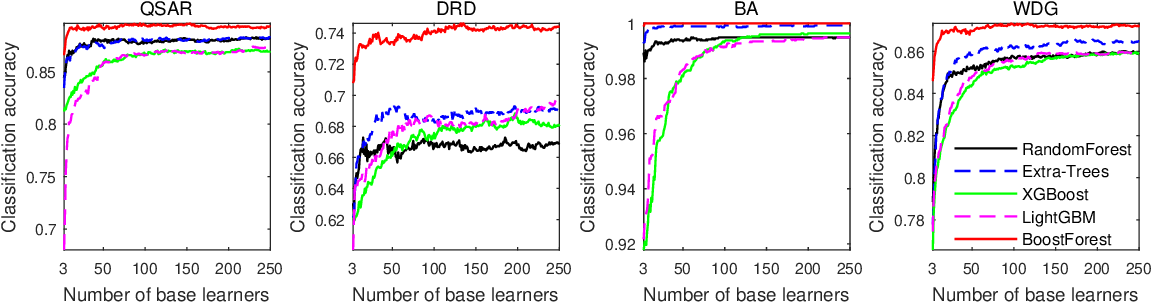}} \\
	\subfigure[]{\label{fig:R2}     \includegraphics[width=.95\linewidth,clip]{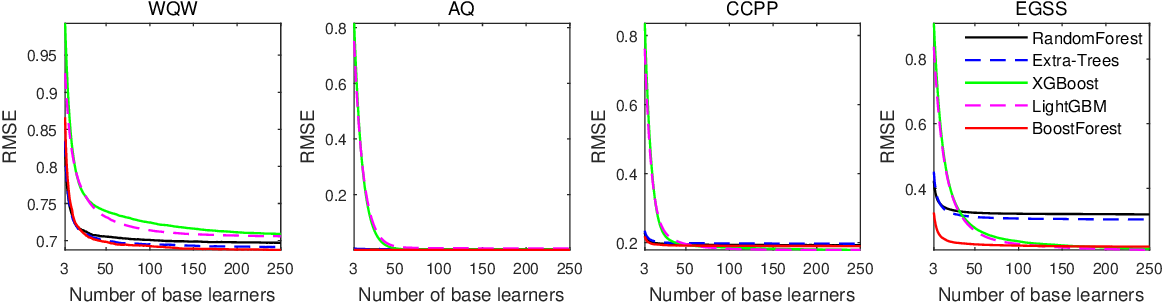}}
	\caption{Generalization performance w.r.t. the number of base learners, averaged over 10 repeats. (a) Average classification accuracies on the last four classification datasets, with different number of base learners. Complete results on the 15 classification datasets are shown in Supplementary Fig.~1. (b) Average RMSEs on the last four regression datasets, with different number of base learners.  Complete results on the 15 regression datasets are shown in Supplementary Fig.~2. }
\end{figure*}

\subsection{Generalization Performance w.r.t. the Base Learner Model Complexity}\label{sec:question3}

We also evaluated the generalization performance of the five ensemble approaches, as the base learner model complexity increases.

The base learner model complexity was controlled by the maximum number of leaves $MaxNumLeaf$ per tree, which was gradually increased from two to 32 for classification, and two to 256 for regression. We fixed the number of base learners at 250, and tuned other parameters of the four baselines by using the validation set (for XGBoost and LightGBM) or the out-of-bag error (for RandomForest and Extra-Trees).

Fig.~\ref{fig:C3} shows the accuracies of the five algorithms on the last four classification datasets. Complete results on all 15 classification datasets are shown in Supplementary Fig.~3. BoostForest achieved the highest classification accuracy on 12 of the 15 datasets.

Fig.~\ref{fig:R3} shows the average RMSEs of the five algorithms on the last four regression datasets. Complete results on all 15 regression datasets are shown in Supplementary Fig.~4. On most datasets, the performances of all algorithms increased as the maximum number of leaves per tree increased. BoostForest achieved the smallest RMSE on seven of the 15 datasets, and the second smallest RMSE on the NO dataset.

These results suggest that BoostForest generalize well with the base learner model complexity.

\begin{figure*} \centering
\subfigure[]{\label{fig:C3}     \includegraphics[width=.95\linewidth,clip]{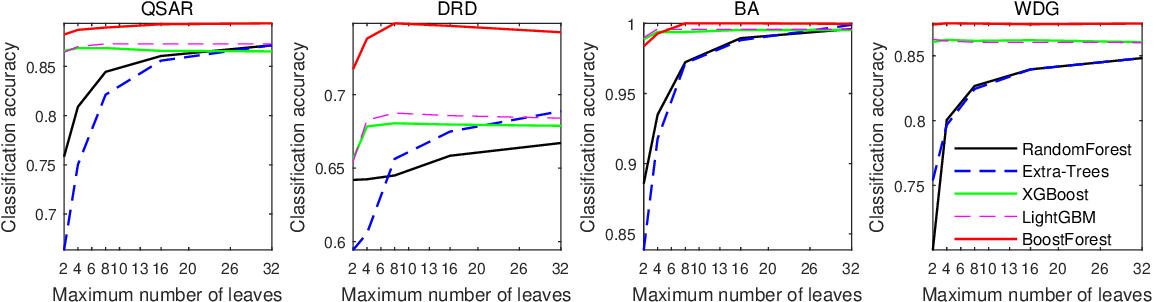}} \\
\subfigure[]{\label{fig:R3}     \includegraphics[width=.95\linewidth,clip]{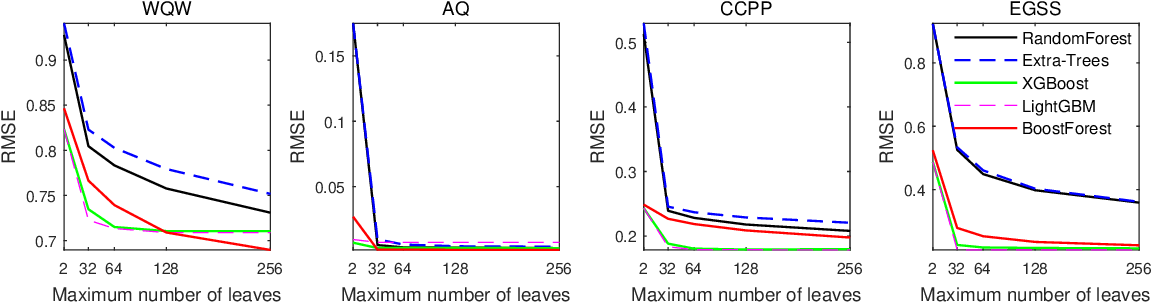}}
\caption{Generalization performance w.r.t. the base learner model complexity, averaged over 10 repeats. (a) Average classification accuracies on the last four classification datasets, with different maximum number of leaves. Complete results on the 15 classification datasets are shown in Supplementary Fig.~3. (b) Average RMSEs on the last four regression datasets, with different maximum number of leaves. Complete results on the 15 regression datasets are shown in Supplementary Fig.~4.} \label{fig:results}
\end{figure*}

\subsection{Generalization Performance on Large Datasets} \label{sec:question4}

Previous experiments have shown the superiority of BoostForest on small to medium sized datasets with not very high dimensionalities. This subsection investigates its performance on four datasets with a large number of samples and/or features.

Table~\ref{tab:bigData} compares the performance of BoostForest with four classical and popular ensemble methods on three classification datasets and two regression datasets. For the SUSY dataset, we set $min\_samples\_leaf=2$ and $MaxNumLeaf=20,000$ in each BoostTree.

BoostForest still demonstrated superior and consistent performance: it achieved the highest average accuracy and the second smallest average time in classification, and the lowest average RMSE and the third smallest average time in regression.

\begin{table*}[htpb]\centering  \setlength{\tabcolsep}{3mm} 
	\caption{Performances of the five ensemble learning approaches on datasets with a large number of samples and/or features. The best performance is marked in bold. $\bullet$ indicates statistically significant win for BoostForest.} \label{tab:bigData}
	\scalebox{0.9}{
		\begin{tabular}{c|llllll}
			\toprule
			\multicolumn{1}{c|}{Dataset}  & RandomForest & Extra-Trees  & XGBoost & LightGBM & BoostForest \\
			\midrule
			&\multicolumn{5}{c}{Mean and standard deviation (in parentheses) of the classification accuracy} \\
			\midrule
			YB               & 0.9913 (0.0033)          & 0.9797 (0.0085)$\bullet$ & 0.9890 (0.0051)          & 0.9874 (0.0049)$\bullet$ & $\textbf{0.9917}$ (0.0023) \\
			LR               & 0.9662 (0.0018)$\bullet$ & 0.9706 (0.0029)$\bullet$ & 0.9637 (0.0026)$\bullet$ & 0.9682 (0.0016)$\bullet$ & $\textbf{0.9763}$ (0.0028) \\
			SUSY             & 0.8026 (0.0004)$\bullet$ & 0.8019 (0.0003)$\bullet$ & 0.8039 (0.0003)$\bullet$ & 0.8038 (0.0004)$\bullet$ & $\textbf{0.8040}$ (0.0003) \\
			\midrule
			Average accuracy & 0.9200 (0.0015)          & 0.9174 (0.0030)          & 0.9189 (0.0023)          & 0.9198 (0.0018)          & $\textbf{0.9240}$ (0.0014) \\
			Average rank     & 3.3333                   & 4.0000                   & 3.3333                   & 3.3333                   & $\textbf{1.0000}$ \\
			Average time                      & 4586.9995 (622.2597)            & \textbf{2514.5877} (580.9703)   & 13698.1652 (2257.4636)               & 5552.3772 (107.2198)                  & 2698.2770 (90.0081)                 \\
			Average model size                      & $3.2834\times10^8$                 &  $5.9279\times10^8$              &  $3.0507\times10^6$                          & $\bm{1.0188\times10^6}$                    & $5.1207\times10^9$                     \\
			\midrule\midrule
			&\multicolumn{5}{c}{Mean and standard deviation (in parentheses) of the regression RMSE} \\
			\midrule
			PTS          & 0.5709 (0.0048)$\bullet$ & 0.5685 (0.0048)$\bullet$   & 0.5713 (0.0050)$\bullet$ & 0.5577 (0.0062)$\bullet$ & $\textbf{0.5504}$ (0.0065) \\
			RLCT         & 0.0617 (0.0054)$\bullet$ & $\textbf{0.0442}$ (0.0030) & 0.0630 (0.0026)$\bullet$ & 0.0605 (0.0026)$\bullet$ & 0.0463 (0.0019)            \\
			\midrule
			Average RMSE & 0.3163 (0.0038)          & 0.3063 (0.0035)            & 0.3172 (0.0027)          & 0.3091 (0.0035)          & $\textbf{0.2983}$ (0.0029) \\
			Average rank & 4.0000                   & 2.0000                     & 5.0000                   & 2.5000                   & $\textbf{1.5000}$ \\
			Average time                      & 183.7901 (2.9807)               & \textbf{97.3998} (2.9675)       & 1550.5950 (551.1156)                 & 1678.8888 (74.8529)                   & 964.5815 (68.9347)                  \\
			Average model size                      & $1.1778\times10^8$                 &  $1.1787\times10^8$             &  $\bm{5.0945\times10^6}$                          & $5.1119\times10^6$                    & $2.5022\times10^9$                     \\
			\bottomrule
	\end{tabular}}%
\end{table*}%

Considering that BoostForest's average model size was the largest and its base learners were more complex than baselines', we also compared BoostForest with Bagging-LightGBM, which uses Bagging to integrate 600 LightGBMs, and BaggedTAO-l, which also uses Bagging to integrate multiple complex trees. To increase BaggedTAO-l's training speed on YB and LR, we selected the cut-point from the set of each feature's \{10, 20, \ldots, 90\} quantiles to initialize each TAO-l in BaggedTAO-l. LMT and M5P are also complex trees, which are compared with BoostForest in Section~\ref{sec:question5}.

Supplementary Table~3 compares BoostForest with Bagging-LightGBM. BoostForest significantly outperformed Bagging-LightGBM on 15 datasets, and achieved higher accuracies on 14 of the 18 classification datasets, and lower RMSEs on 11 of the 17 regression datasets. These results indicated that BoostForest can still achieve promising performance when its average model size was smaller than Bagging-LightGBM.

Supplementary Table~4 compares BoostForest with BaggedTAO-l. BoostForest significantly outperformed BaggedTAO-l on four datasets, and achieved higher accuraies on 16 of the 17 classification datasets. BaggedTAO-l had smaller average model size, because TAO can use the sparsity penalty to remove unnecessary parameters. This idea may also be used to reduce BoostForest's model size.

\subsection{Use Other Base Learners in BoostForest} \label{sec:question5}

Next, we studied if the strategy that BoostForest uses to combine multiple BoostTrees (data replica by bootstrapping, and random parameter selection from a parameter pool) can also be extended to other tree models, i.e., whether we can still achieve good ensemble learning performance when BoostTree is replaced by another tree model.

When Extra-tree, LMT and M5P were used as the tree model, the resulting forests were denoted as ETForest, LMForest and ModelForest, respectively. Individual M5P in ModelForest, LMT in LMForest, and Extra-tree in ETForest, was not pruned, and $min\_samples\_leaf$ was randomly sampled from \{5, 6, \ldots, 15\}. The parameters tuned for the baseline Extra-tree were $max\_depth$ and $min\_samples\_leaf$, whose candidate values were $\{$1, 2, 4, 6, 8, 10$\}$ and $\{$5, 10, 15$\}$, respectively. We set $min\_samples\_leaf=10$ and $\lambda=0.1$ in BoostTree.

Supplementary Table~5 shows the results. ETForest outperformed Extra-tree on 30 of the 32 datasets. ModelForest outperformed M5P on all 16 regression datasets. BoostForest outperformed BoostTree on all 32 datasets. These results indicated that our proposed strategy for integrating BoostTrees into BoostForest can also be used to integrate Extra-tree and M5P into a composite learner with improved performance.

However, LMForest only outperformed LMT on six of the 16 classification datasets, though it achieved better average performance than LMT, i.e., our proposed strategy is not very effective in combining multiple LMTs to further improve their performance. Additionally, using Python to call LMT's and M5P's Java APIs to generate the base learners in parallel is not very efficient, so the average time of LMForest and ModelForest was relatively long.

Finally, BoostForest outperformed LMForest on 14 datasets, among which five were statistically significant; BoostForest outperformed ETForest on 31 datasets, among which 24 were statistically significant; and, BoostForest outperformed ModelForest on 14 datasets, among which 11 were statistically significant.

In summary, BoostForest achieved better average classification performance than LMForest and ETForest, and also better average regression performance than ModelForest and ETForest, indicating that BoostTree is a more effective base learner for our proposed ensemble strategy than Extra-Tree, LMT and M5P.

\subsection{Use Other Regression Models in BoostTree}\label{sec:question6}

We also studied if other more complex and nonlinear regression algorithms, e.g., ELM and SVR, can be used to replace RR as the node function in BoostTree. The resulting trees are denoted as BoostTree-ELM and BoostTree-SVR, respectively, and the corresponding forests as BoostForest-ELM and BoostForest-SVR.

ELM \cite{huang2006extreme} is a single hidden layer neural network. It randomly generates the hidden nodes, and analytically determines the output weights through generalized inverse or RR. Its model complexity can be controlled by the number of hidden nodes \emph{NumHiddenNodes} and the regularization coefficient $\lambda$ of RR. We set their candidate values to $\{$10, 20, 30, 40$\}$ and \{0.0001, 0.001, 0.01, 0.1, 1\}, respectively, to construct the parameter pool. The sigmoid activation function was used in the hidden layer.

Linear SVR \cite{drucker1997support} was used in BoostTree-SVR. The parameter pool for the regularization parameter $C$ and the slack variable $\epsilon$ was $\{$0.01, 0.1, 1.0, 10, 100$\}$ and $\{$0.1, 0.2, 0.4, 0.8, 1.0$\}$, respectively. We set $min\_samples\_leaf=10$ in BoostTree-ELM and BoostTree-SVR, and BoostForest-ELM and BoostForest-SVR to \{5, 6, \ldots, 15\}, respectively. Details of BoostTree-ELM, BoostForest-ELM, BoostTree-SVR and BoostForest-SVR are described in Supplementary Algorithms~2-7.

Supplementary Table~6 compares the performances of ELM and BoostTree-ELM with BoostForest-ELM (SVR and BoostTree-SVR with BoostForest-SVR) on the 15 regression datasets. BoostForest-ELM statistically significantly outperformed ELM (BoostTree-ELM) on 13 (15) datasets. BoostForest-SVR statistically significantly outperformed SVR (BoostTree-SVR) on 14 (15) datasets. BoostTree-ELM and BoostTree-SVR were more likely to overfit, because of their high model complexity. So, it is necessary to combine multiple BoostTrees into BoostForest to reduce over-fitting.

These results showed that when more complex and nonlinear regression algorithms, such as ELM and SVR, are used to replace RR as the node function in BoostTree, the performance may degrade due to overfit; however, integrating the corresponding BoostTrees into a BoostForest can always improve the performance.

We also explored if convolutional neural network (CNN) and deep neural network (DNN) can be used as the node function in BoostForest. The resulting forests are denoted as BoostForest-CNN and BoostForest-DNN, respectively. Supplementary Fig.~5 shows the structures of CNN and DNN used in our experiments. We initialized each node model of BoostForest-CNN or BoostForest-DNN by using its parent node model's weights, scaled its output values to $(-4, 4)$, to be consistent with the range of pseudo-label in (\ref{eq:clip_y}), and trained it with 20 epochs. For the MNIST dataset, we set $min\_samples\_leaf=1,024$, batch size to 1,024, and tree depth to 5 in each BoostTree, and scaled the minimum and maximum values of each feature to 0 and 1, respectively. For the HIGGS dataset, we set $min\_samples\_leaf=2,048$, batch size to 2,048, and tree depth to 5 in each BoostTree, and randomly selected 2,000,000 samples to train each base learner in BoostForest-DNN and Bagging-DNN, respectively. If the number of samples belonging to the splitting node was larger than 100,000, we randomly selected 100,000 samples to identify the cut-point. When training CNN or DNN, we adopted AdamW\footnote{https://pytorch.org/docs/stable/generated/torch.optim.AdamW} with betas of $(0.9, 0.999)$, initial learning rate 0.01, and weight decay 0.0001.

Table~\ref{tab:mnist_higgs} compares the performances of BoostForest-CNN and BoostForest-DNN with six approaches on MNIST and HIGGS, respectively. With less training time, BoostForest-CNN and BoostForest-DNN achieved higher accuracies than Bagging-CNN$_{150}$ and Bagging-DNN$_{150}$, respectively. With smaller model size, BoostForest-CNN (BoostForest-DNN) achieved higher accuracies than RandomForest, Extra-Trees, Bagging-XGBoost, Bagging-lightGBM, and Bagging-CNN$_{3}$ (Bagging-DNN$_{3}$). These results indicated that complex CNN and DNN can also be used as the node function in BoostForest.

\begin{table} \centering
	\caption{Performances of the seven ensemble learning approaches on MNIST and HIGGS. The highest classification accuracy is marked in bold. [$\cdot$] means the training time was recorded by running the approach on a NVIDIA RTX 3090 GPU. CNN$_{150}$ and DNN$_{150}$ were trained with 150 epochs, and CNN$_{3}$ and DNN$_{3}$ were trained with three epochs. } \label{tab:mnist_higgs}
	\scalebox{0.8}{
		\begin{tabular}{c|c|ccc}
			\toprule
			\multicolumn{5}{c}{MNIST} \\ \midrule
			Approach & \makecell[c]{\#Estimators}&  Accuracy          & Time         & Model size         \\ \midrule
			RandomForest & 500 & 0.9724 & $1.7557\times10^2$ & $2.3051\times10^8$ \\
			Extra-Trees  & 500 & 0.9717 & $1.7513\times10^2$ & $3.3620\times10^8$\\
			Bagging-XGBoost   & 500 & 0.9800 & $4.1711\times10^5$ & $2.9380\times10^{9}$\\
			Bagging-LightGBM    & 500 & 0.9817 & $7.2468\times10^4$ & $7.3000\times10^{8}$\\
			Bagging-CNN$_{150}$    & 100 & 0.9871 & [$7.5563\times10^3$] & $1.2560\times10^{6}$\\
			Bagging-CNN$_{3}$    & 6300 & 0.9775 & [$9.3731\times10^3$] & $7.9128\times10^{7}$\\
			BoostForest-CNN    & 100 & \textbf{0.9887}& [$7.4071\times10^3$] & $5.4808\times10^{7}$\\
			\midrule
			\midrule
			\multicolumn{5}{c}{HIGGS} \\ \midrule
			Approach & \makecell[c]{\#Estimators}&  Accuracy          & Time         & Model size         \\ \midrule
			RandomForest & 500 & 0.7633 & $2.6710\times10^4$ & $1.1634\times10^{10}$\\
			Extra-Trees  & 500 & 0.7532 & $2.3760\times10^4$ & $2.7498\times10^{10}$\\
			Bagging-XGBoost   & 500 & 0.7705 & $1.7212\times10^5$ & $1.2034\times10^{10}$\\
			Bagging-LightGBM    & 500 & 0.7740 & $2.8322\times10^5$ & $1.2034\times10^{10}$\\
			Bagging-DNN$_{150}$    & 100 & 0.7873 & [$3.6404\times10^4$] & $1.4009\times10^{8}$\\
			Bagging-DNN$_{3}$    & 6300 & 0.7712 & [$4.6521\times10^4$] & $8.8257\times10^{9}$\\
			BoostForest-DNN    & 100 & \textbf{0.7886} & [$2.6441\times10^4$] & $7.9734\times10^{9}$\\
			\bottomrule
	\end{tabular}}
\end{table}

\subsection{Robustness of BoostForest}\label{sec:question7}

To investigate the robustness of BoostForest, we performed extensive experiments to study how its performance changed with different hyper-parameters. Additionally, we also studied the effect of the clipping operation in (\ref{eq:clip}) and the splitting criterion.

\subsubsection{Effect of $min\_samples\_leaf$ and $\lambda$}\label{Sec:robustness}

There are mainly two hyper-parameters in our proposed BoostForest: $min\_samples\_leaf$ and $\lambda$. Supplementary Table~7 shows the average performances of BoostForest, as $min\_samples\_leaf$ increased from 5 to 15, and $\lambda$ increased from 0.0001 to 1. Randomly selecting each BoostTree's parameters from the parameter pool to form a BoostForest achieved comparable average performance with using the best parameters in both regression and classification, indicating that we can use random parameter pool sampling to simplify BoostForest's parameter selection process.

\subsubsection{Effect of the Clipping Operation in Regression}\label{Sec:clipping}

Clipping makes BoostForest more robust to noise and outliers in regression. To verify the effect of the clipping operation, we compared the performance of $``$BoostForest (w/o clipping)$"$ with $``$BoostForest (w/ clipping)$"$.

Supplementary Table~8 shows the results. BoostForest (w/ clipping) outperformed BoostForest (w/o clipping) on most datasets. When the data noise was small, which means BoostForest can achieve a small RMSE, clipping made BoostForest too conservative to achieve better performance, e.g., on the AQ dataset.

Compared with the baselines in Table~\ref{tab:Ex1}, even BoostForest (w/o clipping) achieved better average RMSE than RandomForest, Extra-Trees, XGBoost, LightGBM and GBDT-PL, indicating that BoostForest (w/o clipping) is also an effective model, though BoostForest (w/ clipping) is more effective.

\subsubsection{Effect of the Splitting Criterion}\label{Sec:Ex_criterion}

We also studied if Gini-SC, C4.5-SC (usually used in C4.5 and LMT) or MSE-SC (usually used in CART and Extra-Trees), can replace XGB-SC in BoostForest.

Supplementary Table~9 shows BoostForest's performances of using different splitting criteria. Using Gini-SC, C4.5-SC and XGB-SC achieved comparable average accuracies in classification. Using XGB-SC achieved better average RMSE than using MSE-SC in regression.

\section{Conclusions and Future Research}\label{sec:conclusion}

This paper has proposed a new tree model, BoostTree, that integrates GBM into a single model tree. BoostTree trains a regression model (for regression or binary classification) or multiple regression models (for multi-class classification) at each node. For a given input, BoostTree first sorts it down to a leaf, then computes the final prediction by summing up the outputs of all node models along the path from the root to that leaf.

Using BoostTrees as base learners, we also proposed a new ensemble learning approach, BoostForest. It first uses bootstrap to obtain multiple replicas of the training set, and then trains a BoostTree on each replica. Its hyper-parameters are easy to tune. Moreover, it represents a very general ensemble learning framework, whose base learners can be any tree model, e.g., BoostTree, Extra-Tree, M5P, or LMT, or even a mixture of different models.

BoostForest performs favorably over several ensemble learning approaches, e.g., RandomForest, Extra-Trees, XGBoost, LightGBM, and GBDT-PL, in both classification and regression, and also MultiBoosting, FilterBoost and DeepBoosting in classification. BoostForest simultaneously uses two randomness injection strategies: 1) data sample manipulation through bootstrapping; and, 2) input feature manipulation through randomly drawing cut-points at node splitting. Output representation manipulation will be considered in our future research.

Zhou and Feng \cite{Zhou2018} showed that Random Forests can be assembled into a Deep Forest to achieve better performance. As we have demonstrated that BoostForest generally outperforms Random Forest, it is also expected that replacing Random Forests in Deep Forest by BoostForests may result in better performance. This is also one of our future research directions.

BoostForest has two main limitations:
\begin{enumerate}
	\item BoostForest cannot handle NULL values, because it needs all features to train a node model.

    \item BoostForest may not be deployed to low storage devices, due to its large model size.
\end{enumerate}
Therefore, our future research will:
\begin{enumerate}
	\item Improve BoostTree to handle NULL values.
	\item Reduce BoostForest's model size by using the sparsity penalty in TAO \cite{TAO_NIPS}.
\end{enumerate}


%

%

\ifCLASSOPTIONcaptionsoff
  \newpage
\fi


\clearpage \onecolumn
\newenvironment{salgorithm}[1][htb]
  {\renewcommand{\algorithmcfname}{Supplementary Algorithm}
   \begin{algorithm}[#1]%
  }{\end{algorithm}}
  
\setcounter{algocf}{0} \setcounter{figure}{0} \setcounter{table}{0}
\renewcommand{\figurename}{Supplementary Fig.}
\renewcommand{\tablename}{Supplementary Table}

\appendices
\section{Supplementary Algorithms}

This section presents pseudo-code of additional algorithms introduced in this paper.

\begin{salgorithm}[!h]
\caption{LogitBoost \cite{friedman2000additive}.}\label{Alg:LogitBoost}
\KwIn{$\{(\bm{x}_n,y_n)\}_{n=1}^N$, $N$ labeled training samples; $K$, the maximum number of iterations; $J$, the number of classes. }
\KwOut{The ensemble $F(\bm{x})$.}
\uIf{$J ==2$ }{
	// $y_n\in \{0,~1\},~n=1,...,N$\;
	Initialize $F(\bm{x})=0$, $w_n=\frac{1}{N}$, and $p(\bm{x}_n)=\frac{1}{2}$, $n=1,2,\dots,N$\;
	\For{$k=1:K$}{
		$\displaystyle z_n =\frac{y_n-p\left(\bm{x}_n\right)}{p\left(\bm{x}_n\right)
			\left[1-p\left(\bm{x}_n\right)\right]}$, $n=1,...,N$\;
		$w_n =p\left(\bm{x}_n\right)\left[1-p\left(\bm{x}_n\right)\right]$, $n=1,...,N$\;
		Fit a function $f_k(\bm{x})$ by weighted least-squares regression, using $\{(\bm{x}_n,z_n)\}_{n=1}^N$ and weights $\{w_n\}_{n=1}^N$\;
		$F(\bm{x})\leftarrow F(\bm{x})+f_k(\bm{x})$\;
		$\displaystyle p(\bm{x}_n)=\frac{1}{1+e^{-F(\bm{x}_n)}}$, $n=1,...,N$\;
	}
}
\Else{
	// $y_n\in \mathbb{R}^{J\times1},~n=1,...,N$\;
	Initialize $F^j(\bm{x})=0$, $w_{n}^j=\frac{1}{N}$, and $p^j(\bm{x}_n)=\frac{1}{J}$, $n=1,2,\dots,N$, $j=1,2,\dots,J$\;
	\For{$k=1:K$}{
		\For{$j=1:J$}{
			$\displaystyle z_n^j =\frac{y_{n}^j-p^j\left(\bm{x}_n\right)}{p^j\left(\bm{x}_n\right)
				\left[1-p^j\left(\bm{x}_n\right)\right]}$, $n=1,...,N$\;
			$w_n^j =p^j\left(\bm{x}_n\right)\left[1-p^j\left(\bm{x}_n\right)\right]$, $n=1,...,N$\;
			Fit a function $f_{k}^j(\bm{x})$ by weighted least-squares regression, using $\{(\bm{x}_n,z_n^j)\}_{n=1}^N$ and weights $\{w_n^j\}_{n=1}^N$\;
		}
		$\displaystyle f_k^j(\bm{x})\leftarrow\frac{J-1}{J}\left[f_k^j
		(\bm{x})-\frac{1}{J}\sum_{i=1}^{J}f_{k}^i(\bm{x})\right]$, $j=1,2,\dots,J$\;
		$F^j(\bm{x})\leftarrow F^j(\bm{x})+f_k^j(\bm{x})$, $j=1,2,\dots,J$\;
		$\displaystyle p^j(\bm{x}_n)=\frac{e^{F^{j}(\bm{x}_n)}}{\sum_{i=1}^Je^{F^i(\bm{x}_n)}}$, $n=1,...,N$, $j=1,2,\dots,J$\;
	}
}
\end{salgorithm}

\begin{salgorithm}[!h]
	\caption{BoostTree-ELM for regression.}\label{alg:BoostTreeELM}
	\KwIn{$\mathrm{Data}=\{(\bm{x}_n,y_n)\}_{n=1}^N$, $N$ training samples, where $\bm{x}_n\in\mathbb{R}^{D\times1}$\;
		\hspace*{10mm} $\mathrm{Pool_{MSL}}$, candidate value pool of the minimum number of samples at a leaf\;
		\hspace*{10mm} $\mathrm{Pool_{NumHiddenNodes}}$, candidate value pool of the number of hidden nodes\;
		\hspace*{10mm} $\mathrm{Pool}_{\lambda}$, candidate value pool of the regularization parameter $\lambda$\;
		\hspace*{10mm} (optional) $\mathrm{MaxNumLeaf}$, the maximum number of leaves, default NULL.}
	\KwOut{A BoostTree-ELM.}
	\vspace*{2mm}
	Randomly select $\lambda$ from $\mathrm{Pool}_{\lambda}$\;
	Randomly select $\mathrm{MSL}$ from $\mathrm{Pool}_\mathrm{MSL}$\;
	Randomly select $M$ from $\mathrm{Pool_{NumHiddenNodes}}$\;
	Initialize $\mathrm{NumLeaf} = 1$, $\mathrm{LeafList} = \{\}$ and $f(\bm{x})=0$\;
	$\mathrm{root}\leftarrow \{\mathrm{data=Data,\ model}=f\}$\;
	Add $\mathrm{root}$ to $\mathrm{LeafList}$\;
	$\mathrm{BoostTree\leftarrow split(root)}$\;
	\vspace*{2mm}
	$\mathrm{split(node)}$\{\\
	Let $m$ be the index of the current node\;
	$\{(\bm{x}_n,y_n)\}_{n\in I_m}\leftarrow \mathrm{node.data}$\;
	$F_m(\bm{x})\leftarrow \sum_{i \in \mathrm{Path}_m}\mathrm{node}_i.\mathrm{model}(\bm{x})$\;
	Initialize $\delta_{\mathrm{gain}}^{\max}=0$ and $\mathrm{split}_\mathrm{flag}=\mathrm{false}$\;
	\For{$d=1:D$}{
		Let $a_{\max}$ and $a_{\min}$ denote the maximal and minimal value of $ \{x_{n,d}|n\in I_m \}$, respectively\;
		Draw a random cut-point $s$ uniformly in $[a_{\min},~a_{\max}]$\;
		$I_L=\{n|x_{n,d}\le s, n\in I_m\}$\;
		$I_R=\{n|x_{n,d}>s, n\in I_m\}$\;
		\If{$|I_L|\geq \mathrm{MSL}$ and $|I_R|\geq \mathrm{MSL}$}
		{
			Calculate $\delta_{\mathrm{gain}}$ in  (13)\;
			\If{$\mathrm{split}_\mathrm{flag}==\mathrm{false}~\mathrm{or}~\delta_{\mathrm{gain}}>\delta_{\mathrm{gain}}^{\max}$}{
				$\delta_{\mathrm{gain}}^{\max}=\delta_{\mathrm{gain}}$, $\mathrm{split}_\mathrm{flag}=\mathrm{true}$, $I_L^*=I_L$, $I_R^*=I_R$\;
			}
		}
	}
	\If{$\mathrm{split}_\mathrm{flag}$ }
	{
		$f_L^*=\mathrm{FitModelELM}(\{(\bm{x}_n,y_n)\}_{n\in I_L^*},\ F_m,\ \lambda,\ M)$\;
		$f_R^*=\mathrm{FitModelELM}(\{(\bm{x}_n,y_n)\}_{n\in I_R^*},\ F_m,\ \lambda,\ M)$\;
		$\mathrm{node.leftChild}=\{\mathrm{data}=\{(\bm{x}_n,y_n)\}_{n\in I_L^*}$, $\mathrm{model}=f_L^*$\}\;
		$\mathrm{node.rightChild=\{data}=\{(\bm{x}_n,y_n)\}_{n\in I_R^*}$, $\mathrm{model}=f_R^*$\}\;
		$\mathrm{NumLeaf=NumLeaf}+1$\;
		Add $\mathrm{node.leftChild}$ and $\mathrm{node.rightChild}$ to $\mathrm{LeafList}$\;
	}
Pop $\mathrm{node}$ from $\mathrm{LeafList}$\;
\If{$|\mathrm{LeafList}|>0$ $ \mathrm{and}$ $\mathrm{(MaxNumLeaf==NULL}$ $\mathrm{or}$ $\mathrm{NumLeaf}< \mathrm{MaxNumLeaf)}$}{
Use (9) to calculate the leaf loss of each $\mathrm{node}$ in $\mathrm{LeafList}$\;
Identify $\mathrm{node^*}$, the leaf node with the highest loss\;
$\mathrm{split(node^*)}$\;
}
\Return node\;
	\}
\end{salgorithm}

\begin{salgorithm}[!h]
	\caption{BoostTree-SVR for regression.}\label{alg:BoostTreeSVR}
	\KwIn{$\mathrm{Data}=\{(\bm{x}_n,y_n)\}_{n=1}^N$, $N$ training samples, where $\bm{x}_n\in\mathbb{R}^{D\times1}$\;
		\hspace*{10mm} $\mathrm{Pool_{MSL}}$, candidate value pool of the minimum number of samples at a leaf\;
		\hspace*{10mm} $\mathrm{Pool}_C$, candidate value pool of the regularization parameter\;
		\hspace*{10mm} $\mathrm{Pool}_{\epsilon}$, candidate value pool of the slack variable\;
		\hspace*{10mm} (optional) $\mathrm{MaxNumLeaf}$, the maximum number of leaves, default NULL.}
	\KwOut{A BoostTree-SVR}
	\vspace*{2mm}
	Randomly select $\lambda$ from $\mathrm{Pool}_{\lambda}$\;
	Randomly select $\mathrm{MSL}$ from $\mathrm{Pool}_\mathrm{MSL}$\;
	Randomly select $C$ from $Pool_{C}$\;
	Randomly select $\epsilon$ from $Pool_{\epsilon}$\;
	Initialize $\mathrm{NumLeaf} = 1$, $\mathrm{LeafList} = \{\}$ and $f(\bm{x})=0$\;
	$\mathrm{root}\leftarrow \{\mathrm{data=Data,\ model}=f\}$\;
	Add $\mathrm{root}$ to $\mathrm{LeafList}$\;
	$\mathrm{BoostTree\leftarrow split(root)}$\;
	\vspace*{2mm}
	$\mathrm{split(node)}$\{\\
	Let $m$ be the index of the current node\;
	$\{(\bm{x}_n,y_n)\}_{n\in I_m}\leftarrow \mathrm{node.data}$\;
	$F_m(\bm{x})\leftarrow \sum_{i \in \mathrm{Path}_m}\mathrm{node}_i.\mathrm{model}(\bm{x})$\;
	Initialize $\delta_{\mathrm{gain}}^{\max}=0$ and $\mathrm{split}_\mathrm{flag}=\mathrm{false}$\;
	\For{$d=1:D$}{
		Let $a_{\max}$ and $a_{\min}$ denote the maximal and minimal value of $ \{x_{n,d}|n\in I_m \}$, respectively\;
		Draw a random cut-point $s$ uniformly in $[a_{\min},~a_{\max}]$\;
		$I_L=\{n|x_{n,d}\le s, n\in I_m\}$\;
		$I_R=\{n|x_{n,d}>s, n\in I_m\}$\;
		\If{$|I_L|\geq \mathrm{MSL}$ and $|I_R|\geq \mathrm{MSL}$}
		{
			Calculate $\delta_{\mathrm{gain}}$ in  (13)\;
			\If{$\mathrm{split}_\mathrm{flag}==\mathrm{false}~\mathrm{or}~\delta_{\mathrm{gain}}>\delta_{\mathrm{gain}}^{\max}$}{
				$\delta_{\mathrm{gain}}^{\max}=\delta_{\mathrm{gain}}$, $\mathrm{split}_\mathrm{flag}=\mathrm{true}$, $I_L^*=I_L$, $I_R^*=I_R$\;
			}
		}
	}
	\If{$\mathrm{split}_\mathrm{flag}$ }
	{
		$f_L^*=\mathrm{FitModelSVR}(\{(\bm{x}_n,y_n)\}_{n\in I_L^*},F_m,C,\epsilon)$\;
		$f_R^*=\mathrm{FitModelSVR}(\{(\bm{x}_n,y_n)\}_{n\in I_R^*},F_m,C,\epsilon)$\;
		$\mathrm{node.leftChild}=\{\mathrm{data}=\{(\bm{x}_n,y_n)\}_{n\in I_L^*}$, $\mathrm{model}=f_L^*$\}\;
		$\mathrm{node.rightChild=\{data}=\{(\bm{x}_n,y_n)\}_{n\in I_R^*}$, $\mathrm{model}=f_R^*$\}\;
		$\mathrm{NumLeaf=NumLeaf}+1$\;
		Add $\mathrm{node.leftChild}$ and $\mathrm{node.rightChild}$ to $\mathrm{LeafList}$\;
	}
Pop $\mathrm{node}$ from $\mathrm{LeafList}$\;
\If{$|\mathrm{LeafList}|>0$ $ \mathrm{and}$ $\mathrm{(MaxNumLeaf==NULL}$ $\mathrm{or}$ $\mathrm{NumLeaf}< \mathrm{MaxNumLeaf)}$}{
Use (9) to calculate the leaf loss of each $\mathrm{node}$ in $\mathrm{LeafList}$\;
Identify $\mathrm{node^*}$, the leaf node with the highest loss\;
$\mathrm{split(node^*)}$\;
}
\Return node\;
	\}
\end{salgorithm}

\begin{salgorithm}[!h]
\caption{$\mathrm{FitModelELM}$ for regression.}\label{Alg:FitELM}
\KwIn{$\{(\bm{x}_n,y_n)\}_{n\in I_c}$, sample set of the current node\;
	\hspace*{10mm} $F_{m}$, ensemble of the models along the path from the root node to the parent node of the current node\;
	\hspace*{10mm} $\lambda$, the regularization parameter\;
	\hspace*{10mm} $M$, the number of hidden nodes.}
\KwOut{ELM model $f_c$ for the current node.}
\vspace*{2mm}
$\tilde{y}_n=y_n-F_{m}(\bm{x}_n)$, $n\in I_c$\;
Fit $f_c = \mathrm{ELM}(\{(\bm{x}_n,\tilde{y}_n)\}_{n\in I_c},\lambda,M)$ using ELM with regularization parameter $\lambda$ and $M$ hidden nodes\;
Clip $f_c$ using (18).
\end{salgorithm}

\begin{salgorithm}[!h]
\caption{BoostForest-ELM for regression.}\label{Alg:ASForestELM}
\KwIn{$\mathrm{Data}=\{(\bm{x}_n,y_n)\}_{n=1}^N$, $N$ training samples\;
\hspace*{10mm} $\mathrm{n\_estimators}$, the number of trees\;
\hspace*{10mm} $\mathrm{Pool_{MSL}}$, candidate value pool of the minimum number of samples at a leaf\;
\hspace*{10mm} $\mathrm{Pool_{NumHiddenNodes}}$, candidate value pool of the number of hidden nodes\;
\hspace*{10mm} $\mathrm{Pool}_{\lambda}$, candidate value pool of the regularization parameter $\lambda$.}
\KwOut{A BoostForest-ELM.}
\vspace*{2mm}
$\mathrm{BoostForest}$-$\mathrm{ELM}=\{\}$\;
\For{$i=1:\mathrm{n\_estimators}$}{
    Bootstrap $\mathrm{Data}'$ from $\mathrm{Data}$\;
    Train $\mathrm{BoostTree}$-$\mathrm{ELM}_i$ on $\mathrm{Data}'$ using Supplementary Algorithm~\ref{alg:BoostTreeELM}\;
    Add $\mathrm{BoostTree}$-$\mathrm{ELM}_i$ to $\mathrm{BoostForest}$-$\mathrm{ELM}$\;
}
\end{salgorithm}

\begin{salgorithm}[!h]
\caption{$\mathrm{FitModelSVR}$ for regression.}\label{Alg:FitSVR}
\KwIn{$\{(\bm{x}_n,y_n)\}_{n\in I_c}$, sample set of the current node\;
	\hspace*{10mm} $F_{m}$, ensemble of the models along the path from the root node to the parent node of the current node\;
	\hspace*{10mm} $C$, the regularization parameter of SVR\;
	\hspace*{10mm} $\epsilon$, the slack variable of SVR.}
\KwOut{The SVR model $f_c$ for the current node.}
\vspace*{2mm}
$\tilde{y}_n=y_n-F_{m}(\bm{x}_n)$, $n\in I_c$\;
Fit $f_c = \mathrm{SVR}(\{\bm{x}_n,\tilde{y}_n\}_{n\in I_c},C,\epsilon)$ using SVR\;
Clip $f_c$ using (18).
\end{salgorithm}

\begin{salgorithm}[!h]
\caption{BoostForest-SVR for regression.}\label{Alg:ASForetSVR}
\KwIn{$\mathrm{Data}=\{(\bm{x}_n,y_n)\}_{n=1}^N$, $N$ training samples\;
\hspace*{10mm} $\mathrm{n\_estimators}$, the number of trees\;
\hspace*{10mm} $\mathrm{Pool_{MSL}}$, candidate value pool of the minimum number of samples at a leaf\;
\hspace*{10mm} $\mathrm{Pool}_\epsilon$, candidate value pool of the slack variable\;
\hspace*{10mm} $\mathrm{Pool}_C$, candidate value pool of the regularization parameter.}
\KwOut{A BoostForest-SVR}
\vspace*{2mm}
$\mathrm{BoostForest}$-$\mathrm{SVR}=\{\}$\;
\For{$i=1:\mathrm{n\_estimators}$}{
    Bootstrap $\mathrm{Data}'$ from $\mathrm{Data}$\;
    Train $\mathrm{BoostTree}$-$\mathrm{SVR}_i$ on $\mathrm{Data}'$ using Supplementary Algorithm~\ref{alg:BoostTreeSVR}\;
    Add $\mathrm{BoostTree}$-$\mathrm{SVR}_i$ to $\mathrm{BoostForest}$-$\mathrm{SVR}$\;
}
\end{salgorithm}

\clearpage
\section{Supplementary Figures}

\begin{figure}[!h] \centering
\includegraphics[width=\linewidth,clip]{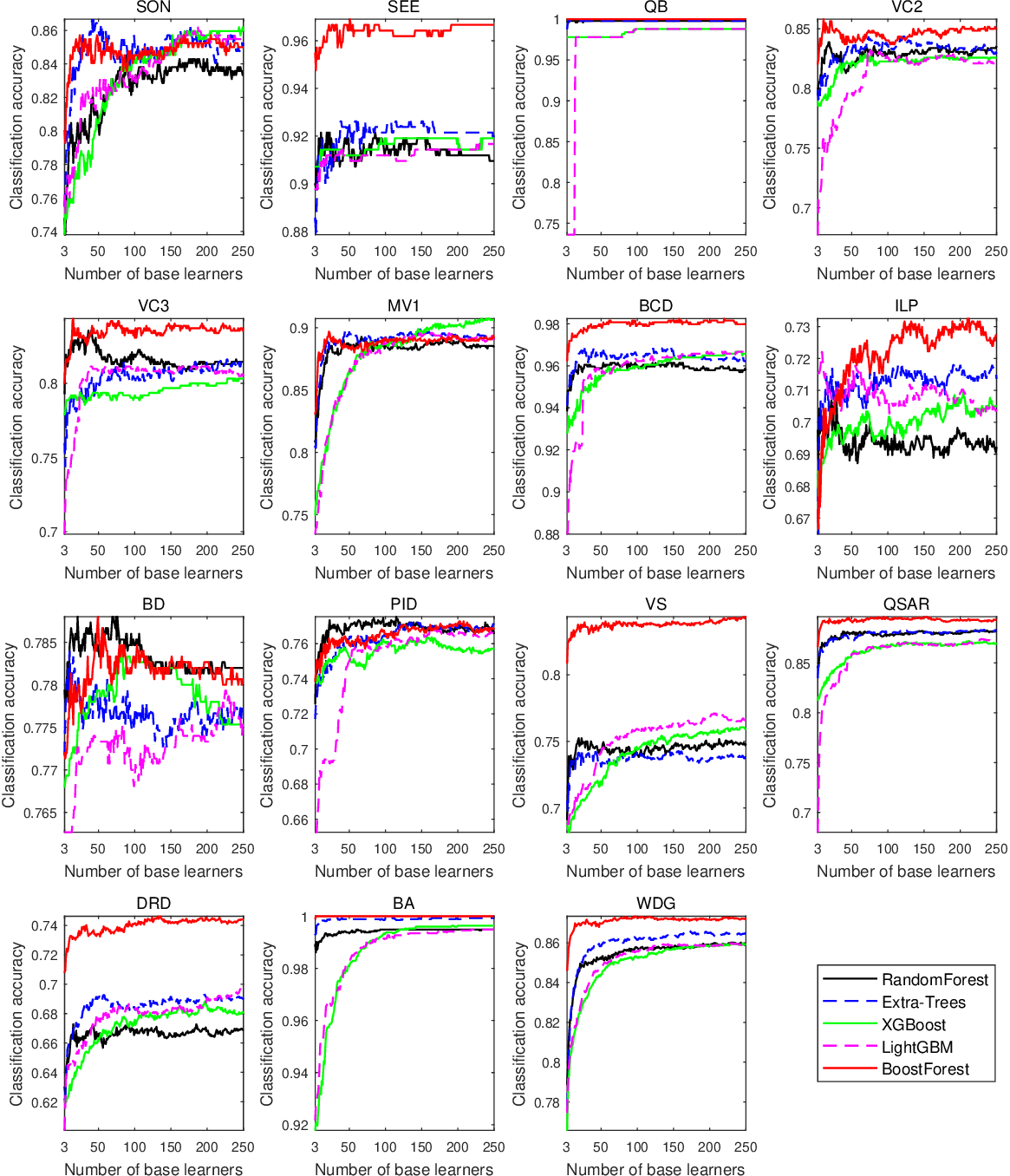}
\caption{Average classification accuracies on the 15 classification datasets, with different number of base learners. } \label{fig:C2}
\end{figure}

\begin{figure}[!h] \centering
\includegraphics[width=\linewidth,clip]{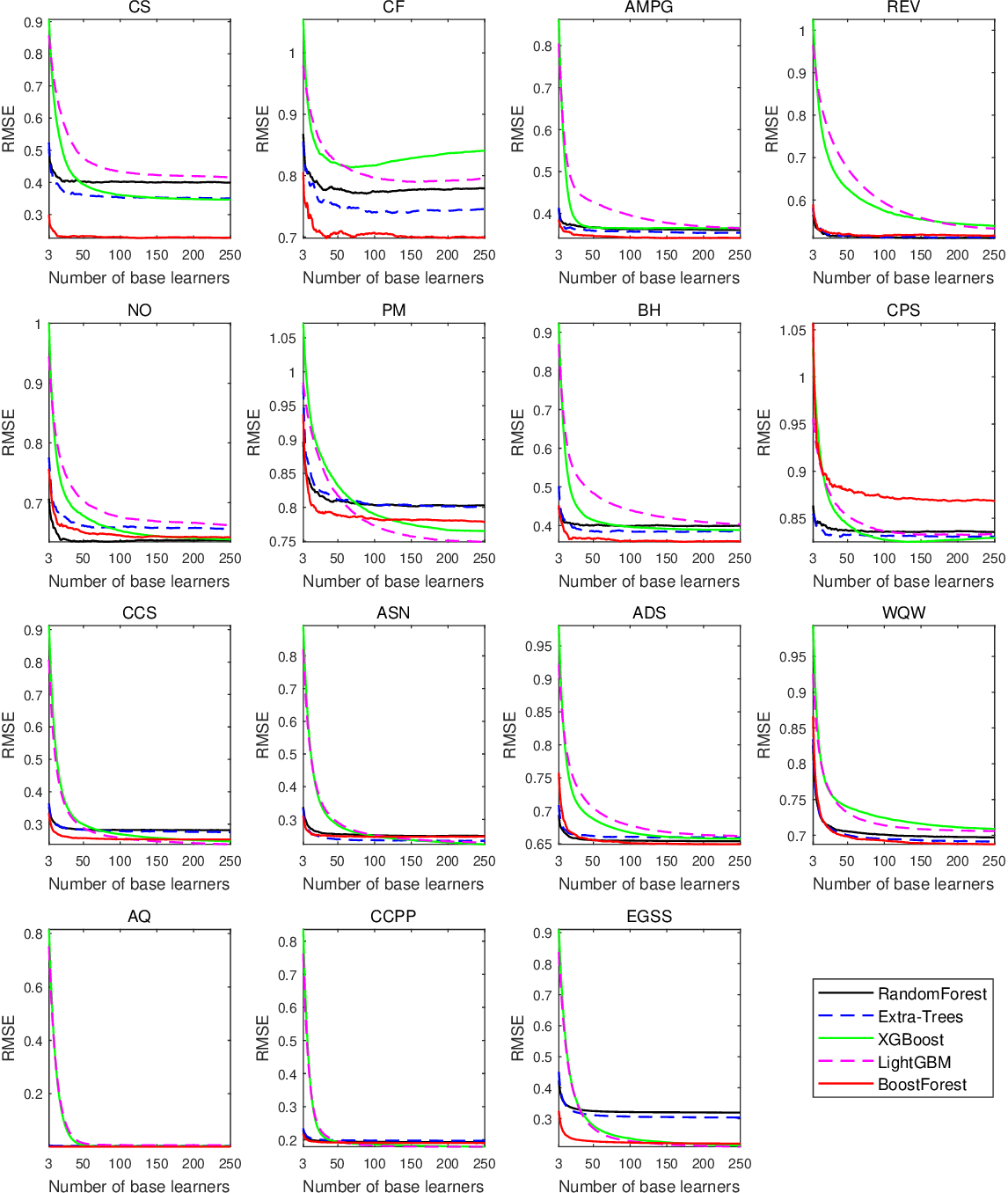}
\caption{Average RMSEs on the 15 regression datasets, with different number of base learners.} \label{fig:R2}
\end{figure}

\begin{figure}[!h] \centering
 \includegraphics[width=\linewidth,clip]{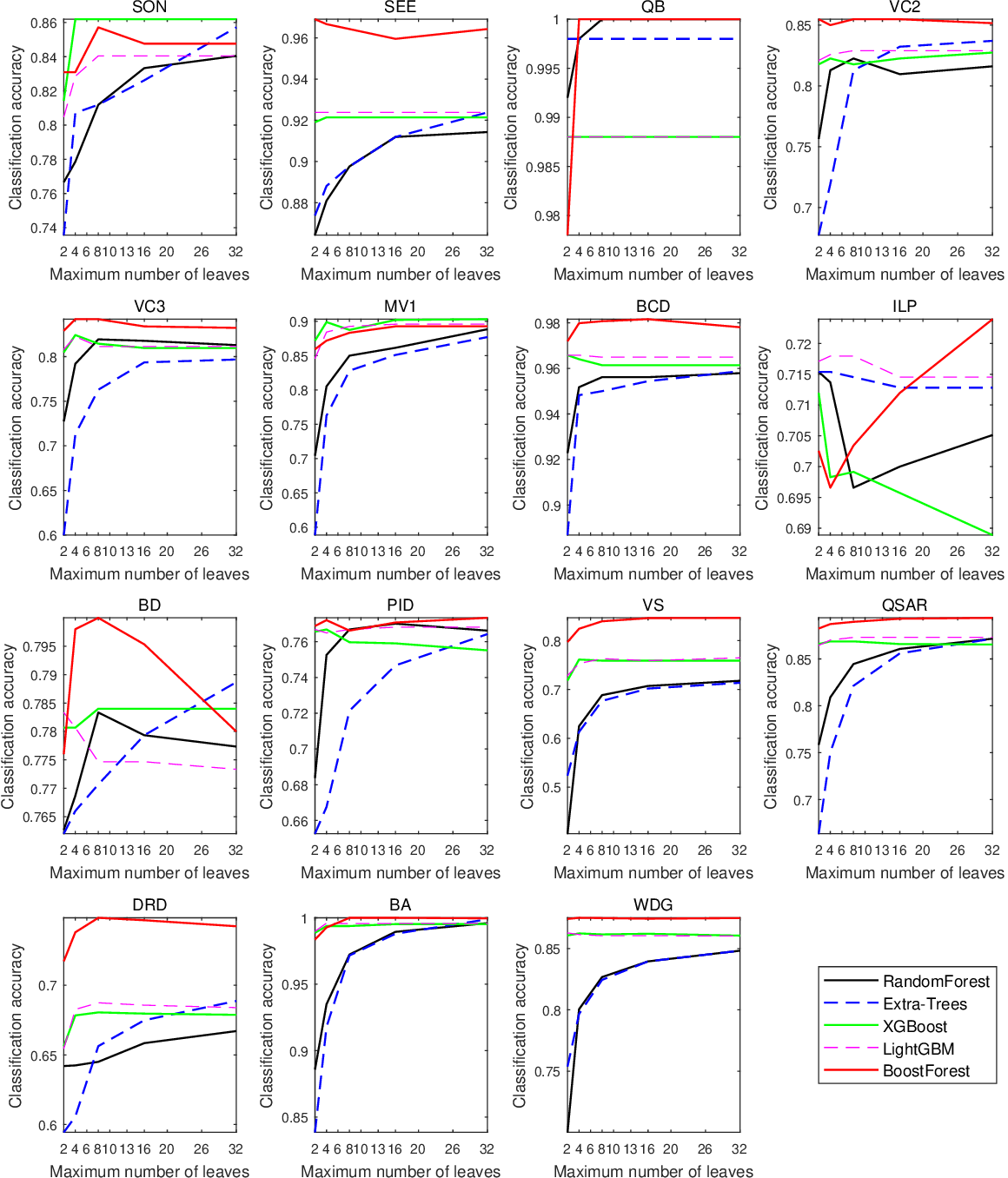}
\caption{Average classification accuracies on the 15 classification datasets, with different maximum number of leaves. } \label{fig:C3}
\end{figure}

\begin{figure}[!h] \centering
 \includegraphics[width=\linewidth,clip]{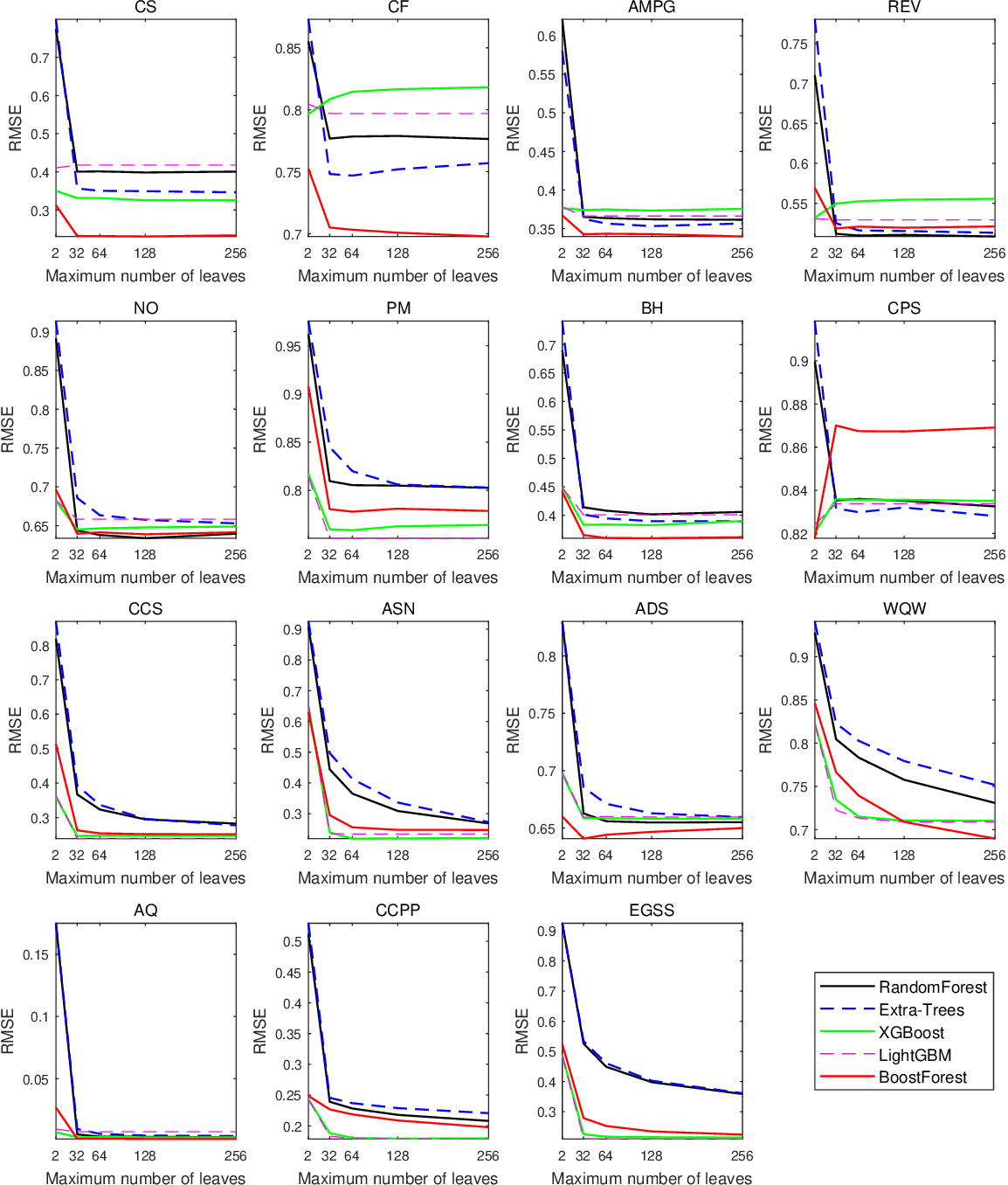}
\caption{Average RMSEs on the 15 regression datasets, with different maximum number of leaves. } \label{fig:R3}
\end{figure}

\begin{figure}[!h] \centering
	\subfigure[]{\label{fig:CNN}     \includegraphics[width=.3\linewidth,clip]{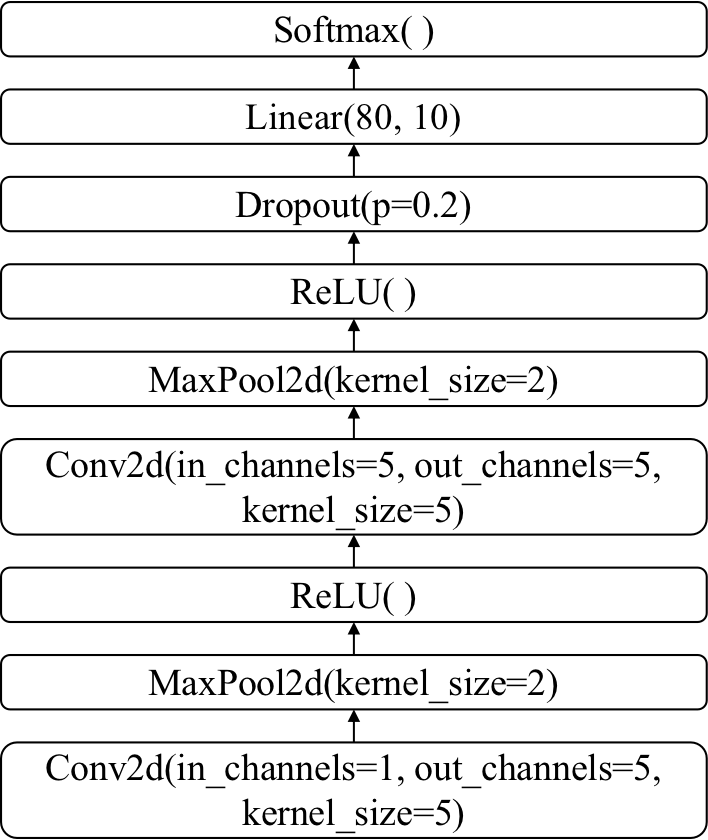}} \\
	\subfigure[]{\label{fig:DNN}     \includegraphics[width=.3\linewidth,clip]{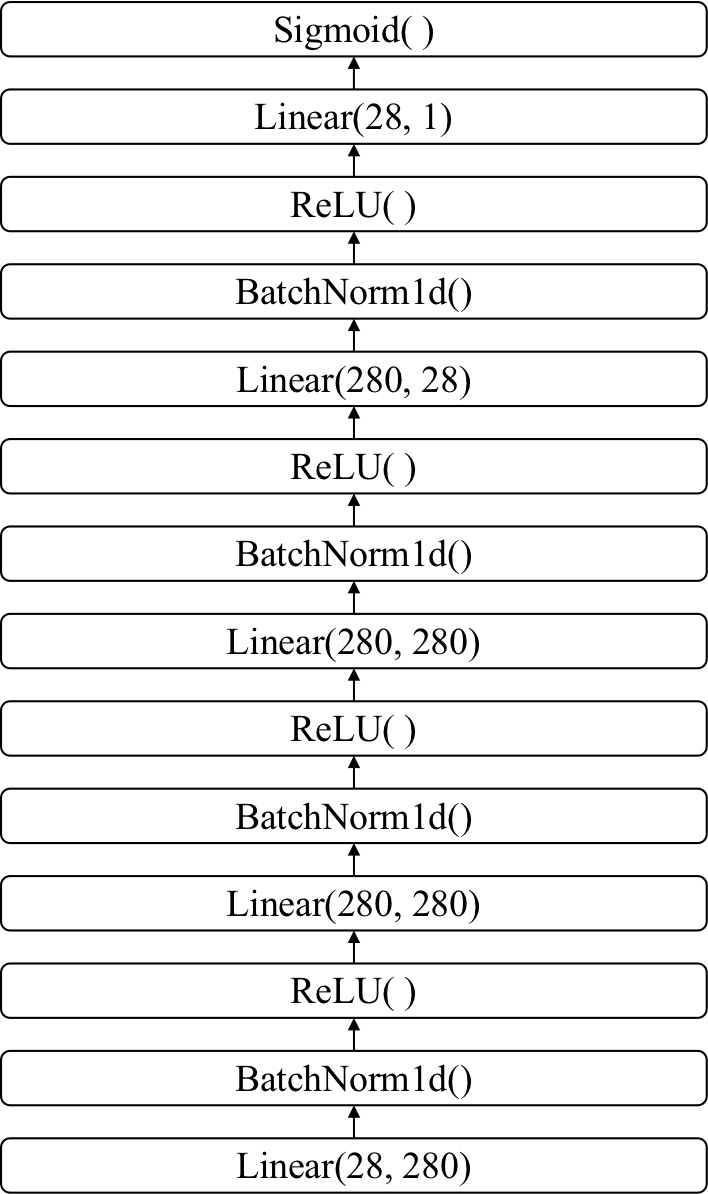}}
	\caption{Structures of neural networks. (a) convolutional neural network (CNN); (b) deep neural network (DNN).} \label{fig:CNNDNN}
\end{figure}

\clearpage
\section{Supplementary Tables}

\begin{table}[htpb] \centering \setlength{\tabcolsep}{5mm} 
	\caption{The 35 datasets used in our experiments.} \label{tab:datasets}
	
	\begin{tabular}{c|llccc}
		\toprule
		Task&Dataset & Abbreviation & \#Samples & \#Features  & \#Classes \\
		\midrule
		&Sonar & SON   & 208   & 60       & 2 \\
		&Seeds & SEE   & 210   & 7        & 3 \\
		&Qualitative Bankruptcy & QB    & 250   & 6        & 2 \\
		&Vertebral Column-2 & VC2   & 310   & 6        & 2 \\
		&Vertebral Column-3 & VC3   & 310   & 6        & 3 \\
		&Musk Version 1 & MV1   & 476   & 166      & 2 \\
		&Breast Cancer Diagnosis & BCD   & 569   & 30       &2 \\
		Classification&Indian Liver Patient & ILP   & 583   & 11       & 2 \\
		&Blood Donations  & BD    & 748   & 4        & 2 \\
		&Pima Indians Diabetes & PID   & 768   & 8        & 2 \\
		&Vehicle Silhouettes & VS    & 846   & 18       & 4 \\
		&QSAR Biodegradation & QSAR  & 1,055  & 41       & 2 \\
		&Diabetic Retinopathy Debrecen & DRD   & 1,151  & 19       & 2 \\
		&Banknote Authentication & BA    & 1,372  & 4        & 2 \\
		&Waveform Database Generator  & WDG   & 5,000  & 21       & 3 \\
		&YaleB  & YB   & 2,414  & 1,024       & 38 \\
		&Letter Recognition  &LR&20,000  & 16       & 26\\
		&MNIST   &MNIST&70,000  & 784       & 10\\
		&SUSY &SUSY&5,000,000  & 18       & 2\\
		&HIGGS &HIGGS&11,000,000  & 28       & 2\\
		\midrule
		&Concrete Slump & CS    & 103   & 7        &  \\
		&Concrete Flow & CF    & 103   & 7        &  \\
		&autoMPG & AMPG  & 392   & 7        &  \\
		&Real Estate Valuation & REV   & 414   & 6        &  \\
		&NO2   & NO    & 500   & 7        &  \\
		&PM10  & PM    & 500   & 7        &  \\
		&Boston Housing & BH    & 506   & 13       &  \\
		Regression&CPS   & CPS   & 534   & 11       &  \\
		&Concrete Compressive Strength & CCS   & 1,030  & 8        &  \\
		&Airfoil Self-Noise & ASN   & 1,503  & 5        &  \\
		&Abalone Data Set & ADS   & 4,177  & 10       &  \\
		&Wine Quality White & WQW   & 4,898  & 11       &  \\
		&Air Quality & AQ    & 9,357  & 8        &  \\
		&Combined Cycle Power Plant & CCPP  & 9,568  & 4        &  \\
		&Electrical Grid Stability Simulated & EGSS  & 10,000 & 12       &  \\
		&Protein Tertiary Structure & PTS  & 45,730 & 9       &  \\
		&Relative Location of CT & RLCT  & 53,500 & 384       &  \\
		\bottomrule
	\end{tabular}
\end{table}%

\begin{table}[htpb]\centering \setlength{\tabcolsep}{0.5mm}
	\caption{Parameter settings for RandomForest, Extra-Trees, XGBoost, LightGBM, GBDT-PL, MultiBoosting, DeepBoosting, FilterBoost, LMT, M5P, BaggedTAO-l and BoostForest. $N$ is the number of samples.} \label{tab:parameters}
	\scalebox{0.97}{
		\begin{tabular}{c|ll}
			\toprule
			Algorithm    & Parameter                                    & Candidate Values                                              \\ \midrule
			\multirow{2}{*}{RandomForest}   & $min\_samples\_leaf$                           & 1, 2, 4, 8                                            \\
			& $n\_estimators$                                & 50, 100, 150, 200, 250                                 \\ \midrule
			\multirow{2}{*}{Extra-Trees}    & $min\_samples\_leaf$                           & 1, 2, 4, 8                                            \\
			& $n\_estimators$                                & 50, 100, 150, 200, 250                                 \\
			& $bootstrap$ & True\\
			\midrule
			\multirow{8}{*}{XGBoost}        & $reg\_alpha$                                   & 0.0001, 0.001, 0.01, 0.1, 1, 10, 100                     \\
			& $reg\_lambda$                                  & 0.0001, 0.001, 0.01, 0.1, 1, 10, 100                     \\
			& $n\_estimators$                                & 250                                                \\
			& $early\_stopping\_rounds$                      & 50                                                 \\
			& $learning\_rate$                               & 0.01, 0.1                                           \\
			& \multirow{3}{*}{$min\_child\_weight$}          & 1, 10, 100 for regression, and                            \\
			&                                              & 0.01, 0.1, 1, 10, 100 for              \\
			&                                              &  classification               \\
			& \multirow{2}{*} {$max\_depth$}                                   & 1, 2, 4, 8 (for $N\le$10,000)                                           \\
			&                                   & 4, 6, 8, 10 (for $N>$10,000)                                           \\
			& \multicolumn{2}{l}{When XGBoost was used as the base learner in} \\
			& \multicolumn{2}{l}{Bagging-XGBoost, the $n\_estimators$ was fixed to 1,000. } \\
			\midrule
			\multirow{8}{*}{LightGBM}       & $reg\_alpha$                                   & 0.0001, 0.001, 0.01, 0.1, 1, 10, 100                     \\
			& $reg\_lambda$                                  & 0.0001, 0.001, 0.01, 0.1, 1, 10, 100                     \\
			& $n\_estimators$                                & 250                                                \\
			& $early\_stopping\_rounds$                      & 50                                                 \\
			& $learning\_rate$                               & 0.01, 0.1                                           \\
			& \multirow{2}{*}{$min\_child\_weight$}          & 1, 10, 100 for regression, and                            \\
			&                                              & 0.01, 0.1, 1, 10, 100 for               \\
			&                                              & classification               \\
			& \multirow{2}{*}{$num\_leaves$ }                                 & $2^{1}$, $2^{2}$, $2^{4}$, $2^{8}$ (for $N\le$10,000)                                           \\
			&                                   & $2^{4}$, $2^{6}$, $2^{8}$, $2^{10}$ (for $N>$10,000)                                           \\
			& \multicolumn{2}{l}{When LightGBM was used as the base learner in} \\
			& \multicolumn{2}{l}{Bagging-LightGBM, the $n\_estimators$ was fixed to 1,000. } \\
			\midrule
			\multirow{11}{*}{GBDT-PL}        & $l2\_reg$                                      & 0.0001, 0.001, 0.01, 0.1, 1, 10, 100                     \\
			& $l1\_reg$                                      & 0.0001, 0.001, 0.01, 0.1, 1, 10, 100                     \\
			& $num\_trees$                                   & 250                                                \\
			& $learning\_rate$                               & 0.01, 0.1                                           \\
			& \multirow{3}{*}{$min\_sum\_hessians$}          & 1, 10, 100 for regression, and                            \\
			&                                              & 0.01, 0.1, 1, 10, 100 for              \\
			&                                              &  classification               \\
			& \multirow{2}{*}{$num\_leaves$ }                                 & $2^{1}$, $2^{2}$, $2^{4}$, $2^{8}$ (for $N\le$10,000)                                           \\
			&                                   & $2^{4}$, $2^{6}$, $2^{8}$, $2^{10}$ (for $N>$10,000)                                           \\
			& $num\_bins$                                    & 255 (followed \cite{gbdtpl2019})        \\
			& $max\_var$                                    & 5  (followed \cite{gbdtpl2019})        \\
			& \multicolumn{2}{l}{The best $num\_leaves$ calculted on the validation set} \\
			& \multicolumn{2}{l}{was used to train the final model and predict the output.} \\
			
			\midrule
			\multirow{2}{*}{MultiBoosting}  & $tree\_depth$                                   & 4, 8, 10, 12                                            \\
			& $n\_estimators$                                & 50, 100, 150, 200, 250                                 \\ \midrule
			\multirow{4}{*}{DeepBoosting}   & $tree\_depth$                                  & 1, 2, 4, 8                                            \\
			& $num\_iter$                                    & 50, 100, 150, 200, 250                                 \\
			&  \multirow{2}{*}{$beta$}                                        & $2^{-6}$, $2^{-5}$, $2^{-4}$, $2^{-3}$, $2^{-2}$, $2^{-1}$, $1$                      \\
			& & (followed \cite{DeepBoosting2014}) \\
			& \multirow{2}{*}{$lambda$}                                        & 0.0001, 0.005, 0.01, 0.05, 0.1, 0.5                \\
			& & (followed \cite{DeepBoosting2014}) \\
			\midrule
			\multirow{3}{*}{FilterBoost}    & $n\_estimators$                                & 250                                                \\
			& $tree\_depth$                                   & 1, 2, 4, 8                                            \\
			& \multicolumn{2}{l}{The best $n\_estimators$ calculted on the validation set} \\
			& \multicolumn{2}{l}{was used to train the final model and predict the output.} \\
			\midrule
			\multirow{4}{*}{BaggedTAO-l}    & $n\_estimators$                                & 250                                                \\
			& $tree\_depth$                                   & 2, 4, 6, 8                                            \\
			& $TAO\_repeat\_iterations$ & 10\\
			& \multicolumn{2}{l}{The best $n\_estimators$ calculted on the validation set} \\
			& \multicolumn{2}{l}{was used to train the final model and predict the output.} \\
			
			\midrule
			LMT            & $-B,~-M,~5$                                &                                                \\ \midrule
			M5P                             & $-M,~ 5$                  & \\
			\midrule
			\multicolumn{3}{c}{Default parameter values of BoostForest}                                                                 \\ \midrule
			\multirow{3}{*}{BoostForest}      & $\mathrm{Pool_{MSL}}$                           & \{5, 6, \ldots, 15\}                                     \\
			& $\mathrm{Pool}_{\lambda}$                                   & \{0.0001, 0.001, 0.01, 0.1, 1\}                      \\
			& $n\_estimators$                                & 250                                                \\
			\bottomrule
	\end{tabular}}
\end{table}

\begin{table}[htpb]\setlength{\tabcolsep}{0.5mm} \centering  \renewcommand\arraystretch{0.9}
	\caption{Performances on the 35 datasets, when BoostForest is compared with Bagging-LightGBM. The best performance is marked in bold. $\bullet$ indicates statistically significant win for BoostForest.}\label{tab:BLGBM_SI}
	\scalebox{1}{
		\begin{tabular}{c|cc}
			\toprule
			\makecell[c]{Classification \\ Dataset} &	\makecell[c]{Bagging- \\ LightGBM}  & BoostForest \\
			\midrule
			&\multicolumn{2}{c}{Mean and standard deviation (in parentheses)}
			\\
			&\multicolumn{2}{c}{of the classification accuracy} \\
			\midrule
			SON              & $\textbf{0.8571}$ (0.0573) & 0.8500 (0.0574)            \\

			SEE              & 0.9357 (0.0239)$\bullet$   & $\textbf{0.9667}$ (0.0117) \\

			QB               & 0.9980 (0.0060)            & $\textbf{1.0000}$ (0.0000) \\

			VC2              & 0.8097 (0.0477)$\bullet$   & $\textbf{0.8500}$ (0.0361) \\

			VC3              & 0.8097 (0.0407)$\bullet$   & $\textbf{0.8371}$ (0.0255) \\

			MV1              & $\textbf{0.8927}$ (0.0291) & 0.8917 (0.0284)            \\

			BCD              & 0.9684 (0.0193)            & $\textbf{0.9798}$ (0.0136) \\

			ILP              & 0.7026 (0.0253)            & $\textbf{0.7274}$ (0.0347) \\

			BD               & 0.7733 (0.0133)            & $\textbf{0.7807}$ (0.0156) \\

			PID              & 0.7617 (0.0277)            & $\textbf{0.7682}$ (0.0197) \\

			VS               & 0.7659 (0.0249)$\bullet$   & $\textbf{0.8429}$ (0.0215) \\

			QSAR             & 0.8773 (0.0226)$\bullet$   & $\textbf{0.8934}$ (0.0225) \\

			DRD              & 0.7052 (0.0265)$\bullet$   & $\textbf{0.7442}$ (0.0192) \\

			BA               & 0.9945 (0.0037)$\bullet$   & $\textbf{1.0000}$ (0.0000) \\

			WDG              & 0.8639 (0.0081)$\bullet$   & $\textbf{0.8716}$ (0.0102) \\

			YB               & $\textbf{0.9925}$ (0.0034) & 0.9917 (0.0023)            \\

			LR               & 0.9681 (0.0029)$\bullet$   & $\textbf{0.9763}$ (0.0028) \\

			SUSY             & $\textbf{0.8042}$ (0.0003) & 0.8040 (0.0003)            \\
\midrule
			Average accuracy & 0.8600 (0.0071)            & $\textbf{0.8764}$ (0.0054) \\

			Average rank     & 1.7778                     & $\textbf{1.2222}$         \\
			Average time & 8921.6813 (853.9904)                    & $\textbf{453.7089}$ (14.9955) \\
			Average model size    &$9.6187\times10^{8}$ &$\bm{8.5823\times10^{8}}$ \\
			\midrule \midrule
			\makecell[c]{Regression \\ Dataset} & \makecell[c]{Bagging- \\ LightGBM}  & BoostForest \\
			\midrule
			&\multicolumn{2}{c}{Mean and standard deviation (in parentheses) } \\
			&\multicolumn{2}{c}{of the regression RMSE} \\
			\midrule
			CS           & 0.3862 (0.0495)$\bullet$   & $\textbf{0.2284}$ (0.0331) \\

			CF           & 0.7983 (0.0879)$\bullet$   & $\textbf{0.6991}$ (0.0721) \\

			AMPG         & 0.3607 (0.0542)            & $\textbf{0.3422}$ (0.0706) \\

			REV          & 0.5208 (0.0825)            & $\textbf{0.5161}$ (0.1005) \\

			NO           & 0.6429 (0.0773)            & $\textbf{0.6422}$ (0.0750) \\

			PM           & $\textbf{0.7483}$ (0.0520) & 0.7786 (0.0427)            \\

			BH           & 0.3866 (0.0809)$\bullet$   & $\textbf{0.3593}$ (0.0705) \\

			CPS          & $\textbf{0.8230}$ (0.0684) & 0.8682 (0.0745)            \\

			CCS          & $\textbf{0.2365}$ (0.0197) & 0.2529 (0.0192)            \\

			ASN          & $\textbf{0.2185}$ (0.0132) & 0.2478 (0.0193)            \\

			ADS          & 0.6517 (0.0262)            & $\textbf{0.6493}$ (0.0187) \\

			WQW          & 0.7018 (0.0302)$\bullet$   & $\textbf{0.6874}$ (0.0324) \\

			AQ           & 0.0074 (0.0030)$\bullet$   & $\textbf{0.0016}$ (0.0013) \\

			CCPP         & $\textbf{0.1805}$ (0.0058) & 0.1902 (0.0062)            \\

			EGSS         & $\textbf{0.1936}$ (0.0026) & 0.2195 (0.0040)            \\

			PTS          & 0.5509 (0.0061)            & $\textbf{0.5504}$ (0.0065) \\

			RLCT         & 0.0552 (0.0028)$\bullet$   & $\textbf{0.0463}$ (0.0019) \\
\midrule
			Average RMSE & 0.4390 (0.0113)            & $\textbf{0.4282}$ (0.0097) \\

			Average rank & 1.6471                     & $\textbf{1.3529}$          \\
			Average time & 1536.6310 (99.3141)                     & \textbf{117.7915} (8.1497) \\
			Average model size    &$1.7997\times10^{9}$ &$\bm{3.0243\times10^{8}}$ \\
			\bottomrule
	\end{tabular}}%
\end{table}%

\begin{table}[htpb]\centering
	\caption{Mean and standard deviation (in parentheses) of the classification accuracy, when BoostForest is compared with BaggedTAO-l. The best performance is marked in bold. $\bullet$ indicates statistically significant win for BoostForest.}\label{tab:TAO_SI}
	\begin{tabular}{@{}ccc@{}}
		\toprule
		\makecell[c]{Classification \\ Dataset}             & BaggedTAO-l                  & BoostForest                    \\ \midrule
		SON              & 0.8429 (0.0429)            & $\textbf{0.8500}$ (0.0574) \\
		SEE              & 0.9476 (0.0233)            & $\textbf{0.9667}$ (0.0117) \\
		QB               & 0.9940 (0.0092)            & $\textbf{1.0000}$ (0.0000) \\
		VC2              & 0.8468 (0.0397)            & $\textbf{0.8500}$ (0.0361) \\
		VC3              & 0.7952 (0.0510)$\bullet$   & $\textbf{0.8371}$ (0.0255) \\
		MV1              & 0.8833 (0.0329)            & $\textbf{0.8917}$ (0.0284) \\
		BCD              & 0.9789 (0.0131)            & $\textbf{0.9798}$ (0.0136) \\
		ILP              & 0.6983 (0.0390)$\bullet$   & $\textbf{0.7274}$ (0.0347) \\
		BD               & $\textbf{0.7833}$ (0.0091) & 0.7807 (0.0156)            \\
		PID              & 0.7636 (0.0191)            & $\textbf{0.7682}$ (0.0197) \\
		VS               & 0.8424 (0.0236)            & $\textbf{0.8429}$ (0.0215) \\
		QSAR             & 0.8806 (0.0272)            & $\textbf{0.8934}$ (0.0225) \\
		DRD              & 0.7368 (0.0220)            & $\textbf{0.7442}$ (0.0192) \\
		BA               & 0.9993 (0.0015)            & $\textbf{1.0000}$ (0.0000) \\
		WDG              & 0.8713 (0.0076)            & $\textbf{0.8716}$ (0.0102) \\
		YB               & 0.9671 (0.0096)$\bullet$   & $\textbf{0.9917}$ (0.0023) \\
		LR               & 0.9604 (0.0028)$\bullet$   & $\textbf{0.9763}$ (0.0028) \\ \midrule
		Average accuracy & 0.8701 (0.0070)            & $\textbf{0.8807}$ (0.0058) \\
		Average rank     & 1.9412                     & $\textbf{1.0588}$          \\
		Average time & 2579.2673 (111.6342)                    & $\textbf{134.2153}$ (15.3546) \\
		Average model size    &$\bm{7.0149\times10^{6}}$ &$8.6048\times10^{8}$ \\ \bottomrule
	\end{tabular}
\end{table}

\begin{table*}[htpb] \centering  \setlength{\tabcolsep}{2mm} 
	\caption{Performances of the eight approaches on the 32 datasets. The best performance is marked in bold. $\bullet$ indicates statistically significant win for BoostForest. } \label{tab:Ex_LMT_M5P_SI}
	\scalebox{1}{
		\begin{tabular}{c|llllll}
			\toprule
			\multicolumn{1}{c|}{Classification Dataset} & LMT & Extra-Tree  & BoostTree & LMForest & ETForest & BoostForest\\
			\midrule
			&\multicolumn{6}{c}{Mean and standard deviation (in parentheses) of the classification accuracy} \\
			\midrule
			SON                               & 0.7690 (0.0674)$\bullet$         & 0.6524 (0.0700)$\bullet$               & 0.7929 (0.0630)$\bullet$                   & $\textbf{0.8500}$ (0.0665)            & 0.8095 (0.0673)                       & 0.8500 (0.0574)                     \\
			SEE                               & 0.9571 (0.0278)                  & 0.7262 (0.1577)$\bullet$               & 0.9524 (0.0261)                            & 0.9405 (0.0220)$\bullet$              & 0.8976 (0.0239)$\bullet$              & $\textbf{0.9667}$ (0.0117)          \\
			QB                                & $\textbf{1.0000}$ (0.0000)       & 0.8860 (0.1368)                        & 0.9980 (0.0060)                            & 0.9980 (0.0060)                       & 0.9980 (0.0060)                       & \textbf{1.0000} (0.0000)                     \\
			VC2                               & 0.8323 (0.0252)                  & 0.6919 (0.0547)$\bullet$               & 0.8242 (0.0284)$\bullet$                   & 0.8371 (0.0310)                       & 0.7597 (0.0441)$\bullet$              & $\textbf{0.8500}$ (0.0361)          \\
			VC3                               & $\textbf{0.8484}$ (0.0501)       & 0.5774 (0.0837)$\bullet$               & 0.7919 (0.0584)$\bullet$                   & 0.8323 (0.0521)                       & 0.7435 (0.0405)$\bullet$              & 0.8371 (0.0255)                     \\
			MV1                               & 0.8427 (0.0356)$\bullet$         & 0.7031 (0.0636)$\bullet$               & 0.8562 (0.0301)$\bullet$                   & $\textbf{0.9083}$ (0.0254)            & 0.8521 (0.0267)$\bullet$              & 0.8917 (0.0284)                     \\
			BCD                               & 0.9754 (0.0140)                  & 0.9079 (0.0465)$\bullet$               & 0.9596 (0.0197)$\bullet$                   & 0.9737 (0.0152)                       & 0.9544 (0.0183)$\bullet$              & $\textbf{0.9798}$ (0.0136)          \\
			ILP                               & 0.7128 (0.0236)                  & 0.7256 (0.0221)                        & 0.6752 (0.0361)$\bullet$                   & 0.7085 (0.0409)                       & 0.7154 (0.0039)                       & $\textbf{0.7274}$ (0.0347)          \\
			BD                                & $\textbf{0.7873}$ (0.0247)       & 0.7660 (0.0125)                        & 0.7767 (0.0180)                            & 0.7800 (0.0235)                       & 0.7653 (0.0058)                       & 0.7807 (0.0156)                     \\
			PID                               & 0.7675 (0.0274)                  & 0.7052 (0.0365)$\bullet$               & 0.7156 (0.0150)$\bullet$                   & 0.7656 (0.0265)                       & 0.7377 (0.0274)$\bullet$              & $\textbf{0.7682}$ (0.0197)          \\
			VS                                & 0.8229 (0.0196)$\bullet$         & 0.5988 (0.0585)$\bullet$               & 0.8059 (0.0258)$\bullet$                   & 0.8035 (0.0220)$\bullet$              & 0.7041 (0.0142)$\bullet$              & $\textbf{0.8429}$ (0.0215)          \\
			QSAR                              & 0.8654 (0.0192)$\bullet$         & 0.7645 (0.0558)$\bullet$               & 0.8555 (0.0274)$\bullet$                   & 0.8834 (0.0207)                       & 0.8720 (0.0290)                       & $\textbf{0.8934}$ (0.0225)          \\
			DRD                               & 0.7134 (0.0348)$\bullet$         & 0.5831 (0.0513)$\bullet$               & 0.6870 (0.0252)$\bullet$                   & 0.7022 (0.0241)$\bullet$              & 0.6831 (0.0272)$\bullet$              & $\textbf{0.7442}$ (0.0192)          \\
			BA                                & 0.9971 (0.0039)                  & 0.9305 (0.0601)$\bullet$               & 0.9967 (0.0044)                            & 0.9989 (0.0017)                       & 0.9967 (0.0041)                       & $\textbf{1.0000}$ (0.0000)          \\
			WDG                               & $\textbf{0.8750}$ (0.0099)       & 0.7169 (0.0229)$\bullet$               & 0.8223 (0.0089)$\bullet$                   & 0.8500 (0.0110)$\bullet$              & 0.8640 (0.0067)$\bullet$              & 0.8716 (0.0102)                     \\
			YB                                & 0.9669 (0.0053)$\bullet$         & 0.2894 (0.0416)$\bullet$               & 0.9526 (0.0152)$\bullet$                   & 0.9878 (0.0035)$\bullet$              & 0.9663 (0.0145)$\bullet$              & $\textbf{0.9917}$ (0.0023)          \\
			\midrule
			Average accuracy                  & 0.8583 (0.0092)                  & 0.7016 (0.0225)                        & 0.8414 (0.0078)                            & 0.8637 (0.0106)                       & 0.8325 (0.0071)                       & $\textbf{0.8747}$ (0.0061)          \\
			Average rank                      & 2.6250                           & 5.6875                                 & 4.3125                                     & 2.9375                                & 4.1875                                & $\textbf{1.2500}$          \\
			Average time                      & 33.7206 (6.0957)                 & \textbf{0.0767} (0.0511)               & 32.9839 (3.5373)                           & 219.9528 (6.1994)                    & 0.4952 (0.0601)                       & 102.9165 (17.1093)                  \\
			Average model size                      & $2.0801\times10^{4}$                 &  $\bm{3.6249\times10^{3}}$             & $2.9305\times10^{6}$                          &  $6.2678\times10^{8}$                    & $1.0452\times10^{6}$                       & $8.4049\times10^{8}$                 \\
			\midrule\midrule
			\multicolumn{1}{c|}{Regression Dataset} & M5P & Extra-Tree  & BoostTree & ModelForest & ETForest & BoostForest\\
			\midrule
			&\multicolumn{6}{c}{Mean and standard deviation (in parentheses) of the regression RMSE} \\
			\midrule
			CS                                & 0.3880 (0.0432)$\bullet$         & 0.7034 (0.1210)$\bullet$               & 0.2543 (0.0633)                            & 0.3536 (0.0510)$\bullet$                 & 0.5908 (0.0793)$\bullet$              & $\textbf{0.2284}$ (0.0331)          \\
			CF                                & 0.8027 (0.0850)$\bullet$         & 0.8968 (0.1372)$\bullet$               & 0.8694 (0.1649)$\bullet$                   & 0.7732 (0.0929)$\bullet$                 & 0.7933 (0.0808)$\bullet$              & $\textbf{0.6991}$ (0.0721)          \\
			AMPG                              & 0.3654 (0.0646)$\bullet$         & 0.4444 (0.0546)$\bullet$               & 0.4480 (0.0985)$\bullet$                   & 0.3538 (0.0586)                          & 0.3727 (0.0665)$\bullet$              & $\textbf{0.3422}$ (0.0706)          \\
			REV                               & 0.5598 (0.0985)$\bullet$         & 0.6446 (0.0931)$\bullet$               & 0.7052 (0.1235)$\bullet$                   & 0.5207 (0.0894)                          & 0.5503 (0.0861)$\bullet$              & $\textbf{0.5161}$ (0.1005)          \\
			NO                                & 0.6831 (0.1009)                  & 0.8341 (0.1010)$\bullet$               & 0.8372 (0.0964)$\bullet$                   & 0.6518 (0.0778)                          & 0.7149 (0.0873)$\bullet$              & $\textbf{0.6422}$ (0.0750)          \\
			PM                                & 0.9157 (0.0481)$\bullet$         & 0.9708 (0.0542)$\bullet$               & 1.0263 (0.0859)$\bullet$                   & 0.8438 (0.0470)$\bullet$                 & 0.8564 (0.0468)$\bullet$              & $\textbf{0.7786}$ (0.0427)          \\
			BH                                & 0.4456 (0.0918)$\bullet$         & 0.5697 (0.0772)$\bullet$               & 0.4988 (0.0830)$\bullet$                   & 0.4256 (0.0942)$\bullet$                 & 0.4575 (0.0953)$\bullet$              & $\textbf{0.3593}$ (0.0705)          \\
			CPS                               & 0.8263 (0.0668)                  & 0.8790 (0.0898)                        & 1.2327 (0.1404)$\bullet$                   & $\textbf{0.8175}$ (0.0585)               & 0.8264 (0.0763)                       & 0.8682 (0.0745)                     \\
			CCS                               & 0.3769 (0.0366)$\bullet$         & 0.5022 (0.0313)$\bullet$               & 0.3593 (0.0266)$\bullet$                   & 0.3333 (0.0197)$\bullet$                 & 0.3787 (0.0192)$\bullet$              & $\textbf{0.2529}$ (0.0192)          \\
			ASN                               & 0.4106 (0.0397)$\bullet$         & 0.5525 (0.0522)$\bullet$               & 0.3380 (0.0421)$\bullet$                   & 0.3660 (0.0190)$\bullet$                 & 0.4366 (0.0261)$\bullet$              & $\textbf{0.2478}$ (0.0193)          \\
			ADS                               & 0.6485 (0.0237)                  & 0.7223 (0.0252)$\bullet$               & 0.8096 (0.0355)$\bullet$                   & $\textbf{0.6432}$ (0.0246)               & 0.6622 (0.0308)                       & 0.6493 (0.0187)                     \\
			WQW                               & 0.8178 (0.0415)$\bullet$         & 0.8662 (0.0295)$\bullet$               & 1.0134 (0.0371)$\bullet$                   & 0.7617 (0.0329)$\bullet$                 & 0.7673 (0.0335)$\bullet$              & $\textbf{0.6874}$ (0.0324)          \\
			AQ                                & 0.1234 (0.0764)$\bullet$         & 0.0213 (0.0089)$\bullet$               & 0.0018 (0.0009)                            & 0.0954 (0.0294)$\bullet$                 & 0.0081 (0.0044)$\bullet$              & $\textbf{0.0016}$ (0.0013)          \\
			CCPP                              & 0.2297 (0.0095)$\bullet$         & 0.2523 (0.0062)$\bullet$               & 0.2279 (0.0085)$\bullet$                   & 0.2108 (0.0056)$\bullet$                 & 0.2184 (0.0053)$\bullet$              & $\textbf{0.1902}$ (0.0062)          \\
			EGSS                              & 0.3578 (0.0073)$\bullet$         & 0.5530 (0.0143)$\bullet$               & 0.4187 (0.0181)$\bullet$                   & 0.3150 (0.0050)$\bullet$                 & 0.3410 (0.0053)$\bullet$              & $\textbf{0.2195}$ (0.0040)          \\
			RLCT                              & 0.1338 (0.0124)$\bullet$         & 0.2717 (0.0117)$\bullet$               & 0.1665 (0.0174)$\bullet$                   & 0.0943 (0.0074)$\bullet$                 & 0.0708 (0.0037)$\bullet$              & $\textbf{0.0463}$ (0.0019)          \\
			\midrule
			Average RMSE                      & 0.5053 (0.0122)                  & 0.6053 (0.0137)                        & 0.5754 (0.0140)                            & 0.4725 (0.0100)                          & 0.5029 (0.0099)                       & $\textbf{0.4205}$ (0.0102)          \\
			Average rank                      & 3.7500                           & 5.4375                                 & 4.6875                                     & 2.2500                                   & 3.5625                                & $\textbf{1.3125}$            \\
			Average time                      & 35.5838 (15.2869)                     & \textbf{0.4621} (0.0385)               & 12.0619 (0.2894)                           & 216.8559 (10.0734)                        & 1.6504 (0.0496)                       & 116.1114 (9.0507)                   \\
			Average model size                      & $1.7820\times10^{5}$                 &  $\bm{2.5703\times10^{3}}$             &  $1.3142\times10^{6}$                          & $5.7983\times10^{7}$                     & $1.9959\times10^6$                       & $3.1193\times10^{8}$                \\
			\bottomrule
	\end{tabular} }
\end{table*}%

\begin{table}\setlength{\tabcolsep}{1mm} \centering  
	\caption{Mean and standard deviation (in parentheses) of the regression RMSE, when ELM or SVR is used to replace RR in BoostTree. The best performance is marked in bold. $\bullet$ indicates statistically significant win for BoostForest-ELM or BoostForest-SVR.}\label{tab:Ex7_SI}
	\scalebox{1}{
		\begin{tabular}{c|lll}
			\toprule
			\makecell[c]{Regression \\ Dataset} & ELM   & BoostTree-ELM & BoostForest-ELM \\
			\midrule
			CS                                & 0.3728 (0.0726)$\bullet$         & 0.3988 (0.1591)$\bullet$             & $\textbf{0.2182}$ (0.0426)           \\
			CF                                & 0.8264 (0.1185)$\bullet$         & 0.9357 (0.1433)$\bullet$             & $\textbf{0.7396}$ (0.0669)           \\
			AMPG                              & 0.3713 (0.0606)$\bullet$         & 0.4507 (0.0993)$\bullet$             & $\textbf{0.3455}$ (0.0716)           \\
			REV                               & 0.6020 (0.0887)$\bullet$         & 0.7902 (0.1397)$\bullet$             & $\textbf{0.5470}$ (0.1017)           \\
			NO                                & 0.7239 (0.0710)$\bullet$         & 0.9228 (0.1311)$\bullet$             & $\textbf{0.6433}$ (0.0691)           \\
			PM                                & 0.9531 (0.0496)$\bullet$         & 1.1628 (0.1831)$\bullet$             & $\textbf{0.8024}$ (0.0427)           \\
			BH                                & 0.5733 (0.0941)$\bullet$         & 0.5640 (0.1350)$\bullet$             & $\textbf{0.3741}$ (0.0724)           \\
			CPS                               & 0.8630 (0.0663)                  & 1.3000 (0.2046)$\bullet$             & $\textbf{0.8579}$ (0.0794)           \\
			CCS                               & 0.5193 (0.0476)$\bullet$         & 0.4070 (0.0641)$\bullet$             & $\textbf{0.2586}$ (0.0229)           \\
			ASN                               & 0.5611 (0.0277)$\bullet$         & 0.3522 (0.0442)$\bullet$             & $\textbf{0.2288}$ (0.0163)           \\
			ADS                               & 0.6600 (0.0254)                  & 0.8698 (0.1131)$\bullet$             & $\textbf{0.6583}$ (0.0219)           \\
			WQW                               & 0.8597 (0.0286)$\bullet$         & 1.0484 (0.0821)$\bullet$             & $\textbf{0.6924}$ (0.0300)           \\
			AQ                                & 0.0362 (0.0068)$\bullet$         & 0.0078 (0.0026)$\bullet$             & $\textbf{0.0037}$ (0.0026)           \\
			CCPP                              & 0.2461 (0.0049)$\bullet$         & 0.2351 (0.0237)$\bullet$             & $\textbf{0.1864}$ (0.0058)           \\
			EGSS                              & 0.6061 (0.0126)$\bullet$         & 0.5616 (0.0662)$\bullet$             & $\textbf{0.2288}$ (0.0043)           \\
			\midrule
			Average RMSE                      & 0.5849 (0.0113)                  & 0.6671 (0.0325)                      & $\textbf{0.4523}$ (0.0117)           \\
			Average rank                      & 2.4000                           & 2.6000                               & $\textbf{1.0000}$ \\
			Average time                      & \textbf{0.1180} (0.0129)                           & 0.5192 (0.0174)                              & 6.5323 (0.1181) \\
			\midrule\midrule
			\makecell[c]{Regression \\ Dataset} & SVR   & BoostTree-SVR & BoostForest-SVR \\
			\midrule
			CS                                & 0.3846 (0.0463)$\bullet$         & 0.5399 (0.1410)$\bullet$             & $\textbf{0.3325}$ (0.0586)           \\
			CF                                & 0.8056 (0.0795)$\bullet$         & 0.8846 (0.0738)$\bullet$             & $\textbf{0.7170}$ (0.0701)           \\
			AMPG                              & 0.4336 (0.0547)$\bullet$         & 0.4211 (0.0785)$\bullet$             & $\textbf{0.3474}$ (0.0675)           \\
			REV                               & 0.6589 (0.0816)$\bullet$         & 0.5998 (0.0993)$\bullet$             & $\textbf{0.5178}$ (0.0955)           \\
			NO                                & 0.7373 (0.0896)$\bullet$         & 0.8121 (0.0992)$\bullet$             & $\textbf{0.6463}$ (0.0728)           \\
			PM                                & 0.9523 (0.0474)$\bullet$         & 1.0419 (0.1373)$\bullet$             & $\textbf{0.7939}$ (0.0428)           \\
			BH                                & 0.5838 (0.1016)$\bullet$         & 0.4708 (0.1368)$\bullet$             & $\textbf{0.3826}$ (0.0699)           \\
			CPS                               & 0.8291 (0.0693)                  & 0.9773 (0.1158)$\bullet$             & $\textbf{0.8290}$ (0.0766)           \\
			CCS                               & 0.6205 (0.0234)$\bullet$         & 0.4053 (0.0378)$\bullet$             & $\textbf{0.2946}$ (0.0163)           \\
			ASN                               & 0.7011 (0.0287)$\bullet$         & 0.4311 (0.0534)$\bullet$             & $\textbf{0.3294}$ (0.0194)           \\
			ADS                               & 0.6841 (0.0245)$\bullet$         & 0.7788 (0.0696)$\bullet$             & $\textbf{0.6438}$ (0.0201)           \\
			WQW                               & 0.8633 (0.0304)$\bullet$         & 0.9278 (0.0825)$\bullet$             & $\textbf{0.7146}$ (0.0330)           \\
			AQ                                & 0.0546 (0.0018)$\bullet$         & 0.0198 (0.0066)$\bullet$             & $\textbf{0.0077}$ (0.0031)           \\
			CCPP                              & 0.2655 (0.0048)$\bullet$         & 0.2383 (0.0176)$\bullet$             & $\textbf{0.2069}$ (0.0052)           \\
			EGSS                              & 0.5976 (0.0089)$\bullet$         & 0.3895 (0.0201)$\bullet$             & $\textbf{0.2442}$ (0.0044)           \\
			\midrule
			Average RMSE                      & 0.6115 (0.0138)                  & 0.5959 (0.0215)                      & $\textbf{0.4672}$ (0.0107)           \\
			Average rank                      & 2.5333                           & 2.4667                               & $\textbf{1.0000}$ \\
			Average time                      & 30.7602 (6.7378)                           & \textbf{1.3324} (0.5250)                               & 40.8591 (3.5997) \\
			\bottomrule
	\end{tabular}}%
\end{table}%
\begin{table} \centering
	\caption{Average performances of BoostForest on the 30 datasets, w.r.t. $min\_samples\_leaf$ (MSL) and $\lambda$. The best performance is marked in bold.}\label{tab:Ex_alpha_MSL}
	\begin{tabular}{l|lll}
		\toprule
		&\multicolumn{3}{c}{Mean and standard deviation (in parentheses)}
		\\
		&\multicolumn{3}{c}{of the classification accuracy} \\
		\midrule
		& MSL=5         & MSL=10        & MSL=15        \\ \midrule
		$\lambda$=0.0001 & 0.8641 (0.0070) & 0.8661 (0.0066) & 0.8658 (0.0070) \\
		$\lambda$=0.001  & 0.8642 (0.0068) & 0.8654 (0.0060) & 0.8663 (0.0073) \\
		$\lambda$=0.01   & 0.8636 (0.0065) & 0.8667 (0.0070) & \textbf{0.8675} (0.0067) \\
		$\lambda$=0.1    & 0.8643 (0.0072) & 0.8665 (0.0061) & 0.8661 (0.0068) \\
		$\lambda$=1.0    & 0.8615 (0.0082) & 0.8630 (0.0078) & 0.8640 (0.0078) \\ \midrule
		\midrule
		&\multicolumn{3}{c}{Mean and standard deviation (in parentheses) } \\
		&\multicolumn{3}{c}{of the regression RMSE} \\
		\midrule
		& MSL=5         & MSL=10        & MSL=15        \\ \midrule
		$\lambda$=0.0001                       & 0.4426 (0.0118)                    & 0.4444 (0.0111)                     & 0.4498 (0.0109)                     \\
		$\lambda$=0.001                        & \textbf{0.4417} (0.0114)                    & 0.4448 (0.0121)                     & 0.4501 (0.0110)                     \\
		$\lambda$=0.01                         & 0.4419 (0.0118)                    & 0.4459 (0.0123)                     & 0.4510 (0.0116)                     \\
		$\lambda$=0.1                          & 0.4441 (0.0122)                    & 0.4472 (0.0099)                     & 0.4535 (0.0100)                     \\
		$\lambda$=1.0                          & 0.4462 (0.0100)                    & 0.4515 (0.0093)                     & 0.4580 (0.0093) \\ \bottomrule
	\end{tabular}
\end{table}
\begin{table}[htpb] \setlength{\tabcolsep}{4mm} \centering  
	\caption{Mean and standard deviation (in parentheses) of the regression RMSE, when BoostForest (w/ clipping) is compared with BoostForest (w/o clipping). The best performance is marked in bold. $\bullet$ indicates statistically significant win for BoostForest (w/ clipping).} \label{tab:Ex_clip_SI}
	\scalebox{1}{
		\begin{tabular}{c|ll}
			\toprule
			\multicolumn{1}{c|}{Regression Dataset} & \multicolumn{1}{c}{\makecell[c]{BoostForest \\ (w/o clipping)}}   & \multicolumn{1}{c}{\makecell[c]{BoostForest \\ (w/ clipping)}} \\
			\midrule
			CS           & 0.2428 (0.0331)            & $\textbf{0.2284}$ (0.0331) \\
			CF           & 0.7319 (0.0782)$\bullet$   & $\textbf{0.6991}$ (0.0721) \\
			AMPG         & 0.3456 (0.0714)            & $\textbf{0.3422}$ (0.0706) \\
			REV          & 0.5407 (0.1125)$\bullet$   & $\textbf{0.5161}$ (0.1005) \\
			NO           & 0.6469 (0.0732)            & $\textbf{0.6422}$ (0.0750) \\
			PM           & 0.7910 (0.0438)            & $\textbf{0.7786}$ (0.0427) \\
			BH           & 0.3887 (0.0704)$\bullet$   & $\textbf{0.3593}$ (0.0705) \\
			CPS          & 0.9003 (0.0753)$\bullet$   & $\textbf{0.8682}$ (0.0745) \\
			CCS          & 0.2681 (0.0182)$\bullet$   & $\textbf{0.2529}$ (0.0192) \\
			ASN          & $\textbf{0.2468}$ (0.0202) & 0.2478 (0.0193)            \\
			ADS          & 0.6531 (0.0238)            & $\textbf{0.6493}$ (0.0187) \\
			WQW          & 0.6986 (0.0301)$\bullet$   & $\textbf{0.6874}$ (0.0324) \\
			AQ           & $\textbf{0.0008}$ (0.0007) & 0.0016 (0.0013)            \\
			CCPP         & 0.1911 (0.0061)            & $\textbf{0.1902}$ (0.0062) \\
			EGSS         & 0.2197 (0.0035)            & $\textbf{0.2195}$ (0.0040) \\
			\midrule
			Average RMSE & 0.4577 (0.0119)            & $\textbf{0.4455}$ (0.0109) \\
			Average rank & 1.8667                     & $\textbf{1.1333}$  \\
			Average time & 4.9382 (0.1262)                     & $\textbf{4.8862}$ (0.1055)  \\
			\bottomrule
	\end{tabular}}%
\end{table}%

\begin{table}[htpb]\setlength{\tabcolsep}{0.5mm} \centering  \renewcommand\arraystretch{0.9}
	\caption{Performances on the 30 datasets, when BoostForest (XGB-SC) is compared with BoostForest (Gini-SC) and BoostForest (MSE-SC). The best performance is marked in bold. $\bullet$ indicates statistically significant win for BoostForest (XGB-SC).}\label{tab:Ex_criterion_SI}
	\scalebox{1}{
		\begin{tabular}{c|lll}
			\toprule
			\makecell[c]{Classification \\ Dataset} & \makecell[c]{BoostForest \\ (Gini-SC)} & \makecell[c]{BoostForest \\ (C4.5-SC)} & \makecell[c]{BoostForest \\ (XGB-SC)} \\
			\midrule
			&\multicolumn{3}{c}{Mean and standard deviation (in parentheses)}
			\\
			&\multicolumn{3}{c}{of the classification accuracy} \\
			\midrule
			SON                               & 0.8500 (0.0593)                      & $\textbf{0.8548}$ (0.0505)          & 0.8500 (0.0574)                     \\
			SEE                               & 0.9619 (0.0117)                      & 0.9524 (0.0238)$\bullet$            & $\textbf{0.9667}$ (0.0117)          \\
			QB                                & $\textbf{1.0000}$ (0.0000)           & \textbf{1.0000} (0.0000)                     & \textbf{1.0000} (0.0000)                     \\
			VC2                               & $\textbf{0.8548}$ (0.0260)           & 0.8323 (0.0347)                     & 0.8500 (0.0361)                     \\
			VC3                               & $\textbf{0.8387}$ (0.0306)           & 0.8097 (0.0344)$\bullet$            & 0.8371 (0.0255)                     \\
			MV1                               & $\textbf{0.8938}$ (0.0286)           & 0.8875 (0.0271)                     & 0.8917 (0.0284)                     \\
			BCD                               & $\textbf{0.9798}$ (0.0136)           & 0.9798 (0.0118)                     & 0.9798 (0.0136)                     \\
			ILP                               & 0.7197 (0.0272)                      & 0.7068 (0.0257)                     & $\textbf{0.7274}$ (0.0347)          \\
			BD                                & 0.7847 (0.0146)                      & $\textbf{0.7900}$ (0.0196)          & 0.7807 (0.0156)                     \\
			PID                               & 0.7636 (0.0218)                      & 0.7545 (0.0153)$\bullet$            & $\textbf{0.7682}$ (0.0197)          \\
			VS                                & $\textbf{0.8476}$ (0.0192)           & 0.8212 (0.0270)                     & 0.8429 (0.0215)                     \\
			QSAR                              & 0.8896 (0.0192)                      & 0.8891 (0.0194)                     & $\textbf{0.8934}$ (0.0225)          \\
			DRD                               & 0.7420 (0.0199)                      & 0.7368 (0.0229)                     & $\textbf{0.7442}$ (0.0192)          \\
			BA                                & $\textbf{1.0000}$ (0.0000)           & 0.9993 (0.0015)                     & \textbf{1.0000} (0.0000)                     \\
			WDG                               & 0.8707 (0.0085)                      & 0.8640 (0.0106)                     & $\textbf{0.8716}$ (0.0102)          \\
			\midrule
			Average accuracy                  & 0.8665 (0.0068)                      & 0.8585 (0.0061)                     & $\textbf{0.8669}$ (0.0064)          \\
			Average rank                      & 1.9333                               & 2.6000                              & $\textbf{1.4667}$ \\
			Average time                      & \textbf{3.7787} (0.0595)                             & 4.4597 (0.0694)                              & 4.7953 (0.0896) \\
			\midrule \midrule
			\makecell[c]{Regression \\ Dataset} & \makecell[c]{BoostForest \\ (MSE-SC)}  & \makecell[c]{BoostForest \\ (XGB-SC)} \\
			\midrule
			&\multicolumn{3}{c}{Mean and standard deviation (in parentheses) } \\
			&\multicolumn{3}{c}{of the regression RMSE} \\
			\midrule
			CS           & 0.2544 (0.0336)$\bullet$   & $\textbf{0.2284}$ (0.0331) \\
			CF           & $\textbf{0.6978}$ (0.0712) & 0.6991 (0.0721)            \\
			AMPG         & 0.3516 (0.0749)$\bullet$   & $\textbf{0.3422}$ (0.0706) \\
			REV          & 0.5208 (0.0989)            & $\textbf{0.5161}$ (0.1005) \\
			NO           & 0.6445 (0.0733)            & $\textbf{0.6422}$ (0.0750) \\
			PM           & 0.8033 (0.0430)$\bullet$   & $\textbf{0.7786}$ (0.0427) \\
			BH           & 0.3705 (0.0639)$\bullet$   & $\textbf{0.3593}$ (0.0705) \\
			CPS          & 0.8767 (0.0878)            & $\textbf{0.8682}$ (0.0745) \\
			CCS          & 0.2660 (0.0232)$\bullet$   & $\textbf{0.2529}$ (0.0192) \\
			ASN          & 0.2693 (0.0174)$\bullet$   & $\textbf{0.2478}$ (0.0193) \\
			ADS          & 0.6532 (0.0192)$\bullet$   & $\textbf{0.6493}$ (0.0187) \\
			WQW          & 0.7018 (0.0365)$\bullet$   & $\textbf{0.6874}$ (0.0324) \\
			AQ           & 0.0017 (0.0010)            & $\textbf{0.0016}$ (0.0013) \\
			CCPP         & 0.2057 (0.0059)$\bullet$   & $\textbf{0.1902}$ (0.0062) \\
			EGSS         & 0.2435 (0.0044)$\bullet$   & $\textbf{0.2195}$ (0.0040) \\
			\midrule
			Average RMSE & 0.4574 (0.0112)            & $\textbf{0.4455}$ (0.0109) \\
			Average rank & 1.9333                     & $\textbf{1.0667}$ \\
			Average time & 6.0168 (0.0819)                    & \textbf{4.8862} (0.1055) \\
			\bottomrule
	\end{tabular}}%
\end{table}%

\end{document}